\documentclass{article}

\usepackage{microtype}
\usepackage{graphicx}
\usepackage{subfigure}
\usepackage{booktabs} 
\usepackage{xcolor}
\usepackage{algorithm}
\usepackage{caption}
\usepackage{hyperref}
\hypersetup{colorlinks=true,breaklinks=true}

\usepackage{amsmath, amsthm, amssymb, multirow, paralist}
\usepackage{mathtools}
\usepackage{bm,bbm}
\usepackage[accepted]{icml2022} 
\usepackage{fontawesome5}
\usepackage{hyperref}

\newtheorem{theorem}{Theorem}
\newtheorem{definition}{Definition}[section]

\def \R {\mathbb{R}}

\icmltitlerunning{Submission and Formatting Instructions for ICML 2022}

\begin{document}

\twocolumn[
\icmltitle{FEDformer: Frequency Enhanced Decomposed Transformer for Long-term Series Forecasting}



\icmlsetsymbol{equal}{*}

\begin{icmlauthorlist}
\icmlauthor{Tian Zhou}{equal,ali}
\icmlauthor{Ziqing Ma}{equal,ali}
\icmlauthor{Qingsong Wen}{ali}
\icmlauthor{Xue Wang}{ali}
\icmlauthor{Liang Sun}{ali}
\icmlauthor{Rong Jin}{ali}
\end{icmlauthorlist}

\icmlaffiliation{ali}{Machine Intelligence Technology, Alibaba Group.}

\icmlcorrespondingauthor{Tian Zhou}{tian.zt@alibaba-inc.com}
\icmlcorrespondingauthor{Rong Jin}{jinrong.jr@alibaba-inc.com}

\icmlkeywords{Machine Learning, ICML}

\vskip 0.3in
]



\printAffiliationsAndNotice{\icmlEqualContribution} 

\begin{abstract}

Although Transformer-based methods have significantly improved state-of-the-art results for long-term series forecasting, they are not only computationally expensive but more importantly, are unable to capture the global view of time series (e.g. overall trend). To address these problems, we propose to combine Transformer with the seasonal-trend decomposition method, in which the decomposition method captures the global profile of time series while Transformers capture more detailed structures. To further enhance the performance of Transformer for long-term prediction, we exploit the fact that most time series tend to have a sparse representation in well-known basis such as Fourier transform, and develop a frequency enhanced Transformer. Besides being more effective, the proposed method, termed as Frequency Enhanced Decomposed Transformer ({\bf FEDformer}), is more efficient than standard Transformer with a linear complexity to the sequence length. Our empirical studies with six benchmark datasets show that compared with state-of-the-art methods, FEDformer can reduce prediction error by $14.8\%$ and $22.6\%$ for multivariate and univariate time series, respectively. Code is publicly available at  https://github.com/MAZiqing/FEDformer.

\end{abstract}



\section{Introduction}\label{sec_intro}
Long-term time series forecasting is a long-standing challenge in various applications (e.g., energy, weather, traffic, economics).
Despite the impressive results achieved by RNN-type methods ~\cite{deep-state-space-models-for-time-series-forecasting,DBLP:journals/corr/FlunkertSG17-deepAR}, they often suffer from the problem of gradient vanishing or exploding ~\cite{DBLP:icml/On-the-difficult-gradient-vanishing-explode}, significantly limiting their performance. Following the recent success in NLP and CV community~\cite{attention_is_all_you_need,Bert/NAACL/Jacob,Transformers-for-image-at-scale/iclr/DosovitskiyB0WZ21,DBLP:Global-filter-FNO-in-cv}, Transformer ~\cite{attention_is_all_you_need} has been introduced to capture long-term dependencies in time series forecasting and shows promising results~\cite{haoyietal-informer-2021,Autoformer}. Since high computational complexity and memory requirement make it difficult for Transformer to be applied to long sequence modeling, numerous studies are devoted to reduce the computational cost of Transformer~\cite{Log-transformer-shiyang-2019,DBLP:conf/iclr/KitaevKL20-reformer,haoyietal-informer-2021,DBLP:journals/corr/abs-2006-04768-linformer,DBLP:conf/aaai/XiongZCTFLS21/Nystroformer,DBLP:journals/corr/LUNA}. A through overview of this line of works can be found in Appendix \ref{App:related_works}.

\begin{figure}[h]
\begin{minipage}{\linewidth/2}
    \centering
    \scalebox{0.95}{
    \includegraphics[width=1\linewidth]{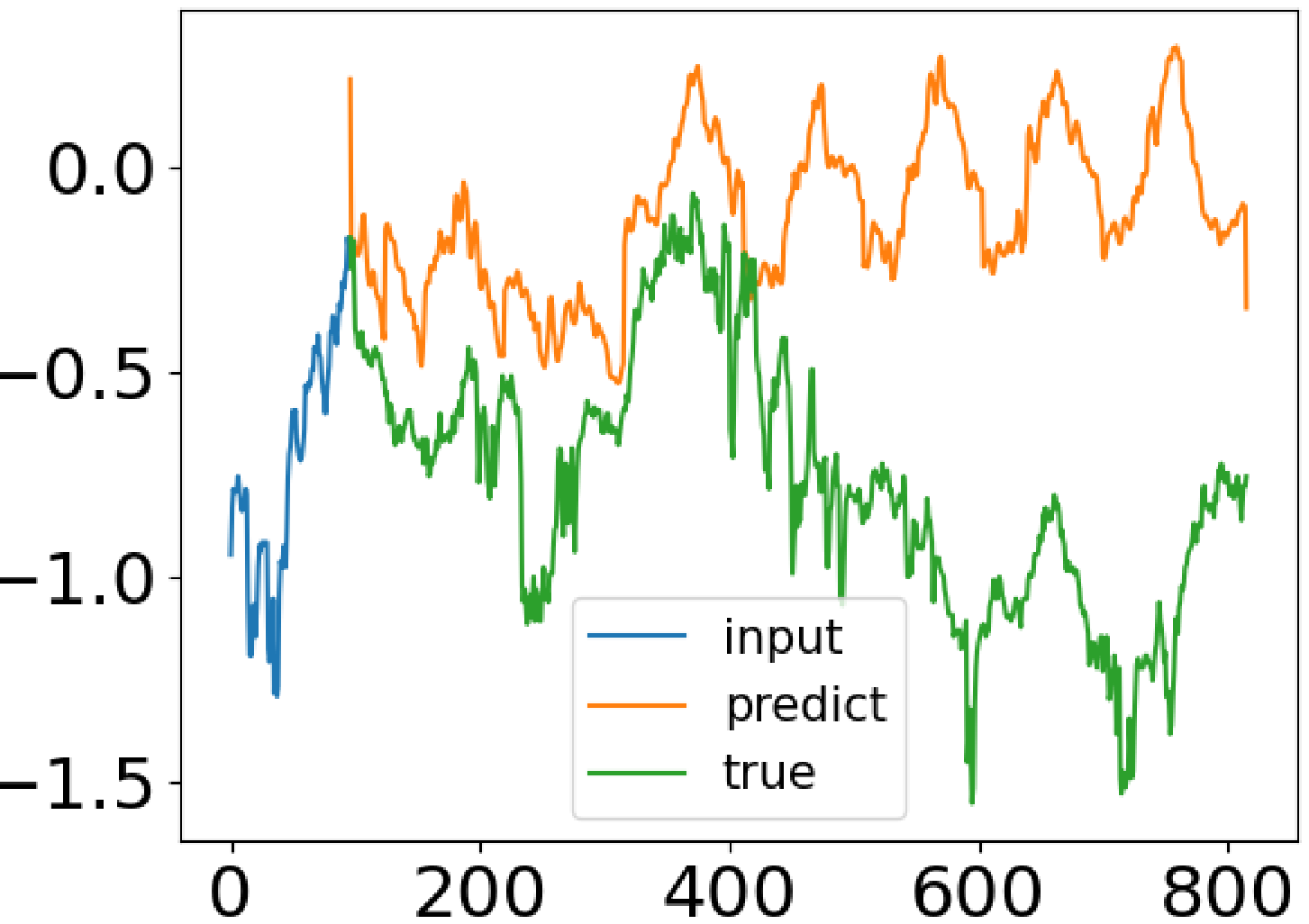}}
    \end{minipage}\hfill
    \begin{minipage}{\linewidth/2}
     \centering
     \scalebox{0.95}{
    \includegraphics[width=1\linewidth]{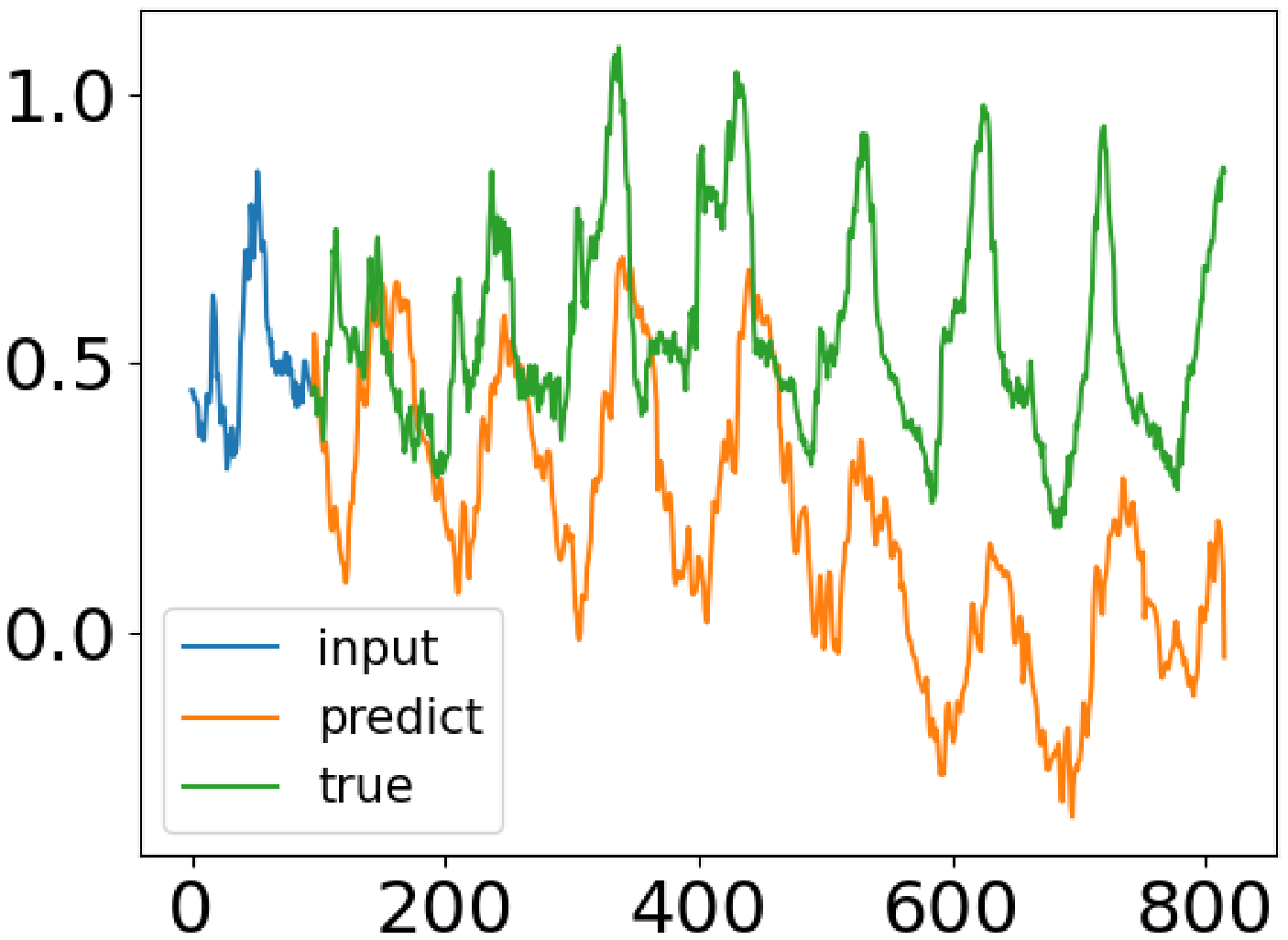}}
    \end{minipage}\hfill
    \caption{Different distribution between ground truth and forecasting output from vanilla Transformer in a real-world ETTm1 dataset. Left: frequency mode and trend shift. Right: trend shift.}
    \label{fig:output_dis_analy_0}
    \vskip -0.15in
\end{figure}

Despite the progress made by Transformer-based methods for time series forecasting, they tend to fail in capturing the overall characteristics/distribution of time series in some cases. In Figure \ref{fig:output_dis_analy_0}, we compare the time series of ground truth with that predicted by the vanilla Transformer method~\cite{attention_is_all_you_need} in a real-world ETTm1 dataset~\cite{haoyietal-informer-2021}. It is clear that the predicted time series shared a {\it different} distribution from that of ground truth. The discrepancy between ground truth and prediction could be explained by the point-wise attention and prediction in Transformer. Since prediction for each timestep is made individually and independently, it is likely that the model fails to maintain the global property and statistics of time series as a whole. To address this problem, we exploit two ideas in this work. The first idea is to incorporate a seasonal-trend decomposition approach~\cite{cleveland1990stl,wen2019robuststl}, which is widely used in time series analysis, into the Transformer-based method. Although this idea has been exploited before~\cite{nbeats, Autoformer}, we present a special design of network that is effective in bringing the distribution of prediction close to that of ground truth, according to Kologrov-Smirnov distribution test. Our second idea is to combine Fourier analysis with the Transformer-based method. Instead of applying Transformer to the time domain, we apply it to the frequency domain which helps Transformer better capture global properties of time series. Combining both ideas, we propose a \textbf{F}requency \textbf{E}nhanced \textbf{D}ecomposition Trans\textbf{former}, or, \textbf{FEDformer} for short, for long-term time series forecasting. 

One critical question with FEDformer is which subset of frequency components should be used by Fourier analysis to represent time series. A common wisdom is to keep low-frequency components and throw away the high-frequency ones. This may not be appropriate for time series forecasting as some of trend changes in time series are related to important events, and this piece of information could be lost if we simply remove all high-frequency components. We address this problem by effectively exploiting the fact that time series tend to have (unknown) sparse representations on a basis like Fourier basis. According to our theoretical analysis, a randomly selected subset of frequency components, including both low and high ones, will give a better representation for time series, which is further verified by extensive empirical studies. Besides being more effective for long term forecasting, combining Transformer with frequency analysis allows us to reduce the computational cost of Transformer from quadratic to linear complexity. We note that this is different from previous efforts on speeding up Transformer, which often leads to a performance drop.

In short, we summarize the key contributions of this work as follows:
\begin{enumerate}
    \item We propose a {\it frequency enhanced decomposed Transformer} architecture with mixture of experts for seasonal-trend decomposition in order to better capture global properties of time series.
    \item We propose {\it Fourier enhanced blocks} and {\it Wavelet enhanced blocks} in the Transformer structure that allows us to capture important structures in time series through frequency domain mapping. They serve as substitutions for both self-attention and cross-attention blocks. 
    \item By randomly selecting a fixed number of Fourier components, the proposed model achieves {\it linear computational complexity and memory cost}. The effectiveness of this selection method is verified both theoretically and empirically.

    \item We conduct extensive experiments over 6 benchmark datasets across multiple domains (energy, traffic, economics, weather and disease). Our empirical studies show that the proposed model improves the performance of state-of-the-art methods by $14.8\%$ and $22.6\%$ for multivariate and univariate forecasting, respectively. 
\end{enumerate}

%

\begin{figure*}[htbp]
\centering
\scalebox{0.95}{
\includegraphics[width=\linewidth]{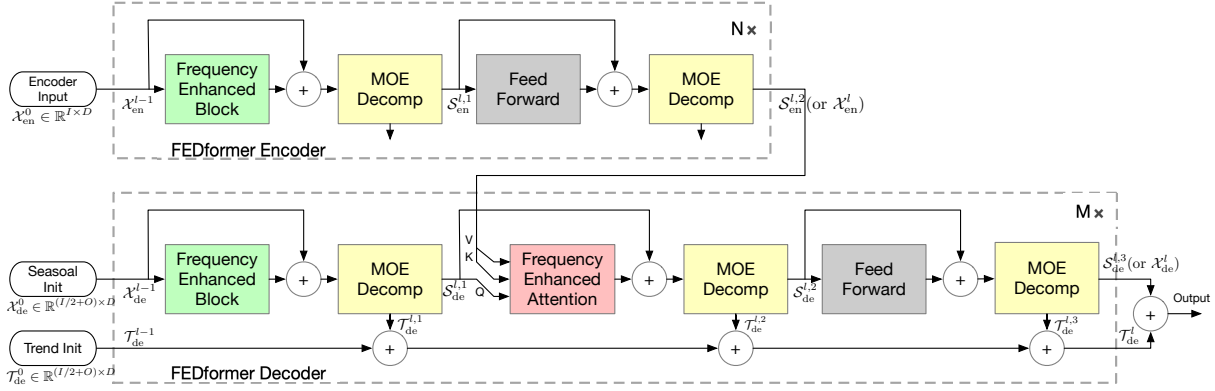}}
\caption{FEDformer Structure. The FEDformer consists of $N$ encoders and $M$ decoders. The Frequency Enhanced Block (FEB, \textcolor{green}{green} blocks) and Frequency Enhanced Attention (FEA, \textcolor{red}{red} blocks) are used to perform representation learning in frequency domain. Either FEB or FEA has two subversions (FEB-f \& FEB-w or FEA-f \& FEA-w), where `-f' means using Fourier basis and `-w' means using Wavelet basis. The Mixture Of Expert Decomposition Blocks (MOEDecomp, \textcolor{yellow}{yellow} blocks) are used to extract seasonal-trend patterns from the input data.}
\label{fig:FEDformerStructure}
\vskip -0.15in
\end{figure*}

\section{Compact Representation of Time Series in Frequency Domain}
\label{sec:signal_represent_random_f_comp}
It is well-known that time series data can be modeled from the time domain and frequency domain. One key contribution of our work which separates from other long-term forecasting algorithms is the frequency-domain operation with a neural network. As Fourier analysis is a common tool to dive into the frequency domain, while how to appropriately represent the information in time series using Fourier analysis is critical. Simply keeping all the frequency components may result in inferior representations since many high-frequency changes in time series are due to noisy inputs. On the other hand, only keeping the low-frequency components may also be inappropriate for series forecasting as some trend changes in time series represent important events. Instead, keeping a compact representation of time series using a small number of selected Fourier components will lead to efficient computation of transformer, which is crucial for modelling long sequences. We propose to represent time series by randomly selecting a constant number of Fourier components, including both high-frequency and low-frequency. Below, an analysis that justifies the random selection is presented theoretically. Empirical verification can be found in the experimental session. 

Consider we have $m$ time series, denoted as $X_1(t), \ldots, X_m(t)$. By applying Fourier transform to each time series, we turn each $X_i(t)$ into a vector $a_i = (a_{i,1}, \ldots, a_{i,d})^{\top} \in \R^d$. By putting all the Fourier transform vectors into a matrix, we have $A = (a_1, a_2, \ldots, a_m)^{\top} \in \R^{m\times d}$, with each row corresponding to a different time series and each column corresponding to a different Fourier component. Although using all the Fourier components allows us to best preserve the history information in the time series, it may potentially lead to overfitting of the history data and consequentially a poor prediction of future signals. Hence, we need to select a subset of Fourier components, that on the one hand should be small enough to avoid the overfitting problem and on the other hand, should be able to preserve most of the history information. Here, we propose to select $s$ components from the $d$ Fourier components ($s < d$) uniformly at random. More specifically, we denote by $i_1 < i_2 < \ldots < i_s$ the randomly selected components. We construct matrix $S \in \{0, 1\}^{s\times d}$, with $S_{i, k} = 1$ if $i = i_k$ and $S_{i, k} = 0$ otherwise. Then, our representation of multivariate time series becomes $A' = AS^{\top} \in \R^{m\times s}$. Below, we will show that, although the Fourier basis are randomly selected, under a mild condition, $A'$ is able to preserve most of the information from $A$. 

In order to measure how well $A'$ is able to preserve information from $A$, we project each column vector of $A$ into the subspace spanned by the column vectors in $A'$. We denote by $P_{A'}(A)$ the resulting matrix after the projection, where $P_{A'}(\cdot)$ represents the projection operator. If $A'$ preserves a large portion of information from $A$, we would expect a small error between $A$ and $P_{A'}(A)$, i.e. $|A - P_{A'}(A)|$. 
Let $A_k$ represent the approximation of $A$ by its first $k$ largest single value decomposition. 
The theorem below shows that $|A - P_{A'}(A)|$ is close to $|A - A_k|$ if the number of randomly sampled Fourier components $s $ is on the order of $k^2$. 
\begin{theorem}\label{theorem1}
Assume that $\mu(A)$, the coherence measure of matrix $A$, is $\Omega(k/n)$. Then, with a high probability, we have
\[
    |A - P_{A'}(A)| \leq (1 + \epsilon)|A - A_k|
\]
if $s = O(k^2/\epsilon^2)$.
\end{theorem} 
The detailed analysis can be found in Appendix  \ref{app:Fourier_component_selection}. 

For real-world multivariate times series, the corresponding matrix $A$ from Fourier transform often exhibit low rank property, since those univaraite variables in multivariate times series depend not only on its past values but also has dependency on each other, as well as share similar frequency components. Therefore, as indicated by the Theorem~\ref{theorem1}, randomly selecting a subset of Fourier components allows us to appropriately represent the information in Fourier matrix $A$.

Similarly, wavelet orthogonal polynomials, such as Legendre Polynomials, obey restricted isometry property (RIP) and can be used for capture information in time series as well. Compared to Fourier basis, wavelet based representation is more effective in capturing local structures in time series and thus can be more effective for some forecasting tasks. We defer the discussion of wavelet based representation in Appendix \ref{app:Low_rank_approximation}. In the next section, we will present the design of frequency enhanced decomposed Transformer architecture that incorporate the Fourier transform into transformer.



\section{Model Structure}
In this section, we will introduce (1) the overall structure of FEDformer, as shown in Figure~\ref{fig:FEDformerStructure}, (2) two subversion structures for signal process: one uses Fourier basis and the other uses Wavelet basis, (3) the mixture of experts mechanism for seasonal-trend decomposition, and (4) the complexity analysis of the proposed model.

\subsection{FEDformer Framework}
\paragraph{Preliminary}
Long-term time series forecasting is a sequence to sequence problem. We denote the input length as $I$ and output length as $O$. We denote $D$ as the hidden states of the series. The input of the encoder is a $I\times D$ matrix and the decoder has $(I/2+O) \times D$ input.

\paragraph{FEDformer Structure}
Inspired by the seasonal-trend decomposition and distribution analysis as discussed in Section~\ref{sec_intro}, we renovate Transformer as a deep decomposition architecture as shown in Figure \ref{fig:FEDformerStructure}, including Frequency Enhanced Block (FEB), Frequency Enhanced Attention (FEA) connecting encoder and decoder, and the Mixture Of Experts Decomposition block (MOEDecomp). The detailed description of FEB, FEA, and MOEDecomp blocks will be given in the following Section~\ref{sec_Fourierblock}, \ref{sec_Waveletblock}, and \ref{sec_MOEDecomp} respectively. 

The encoder adopts a multilayer structure as: $\mathcal{X}_{\mathrm{en}}^{l}=\text{Encoder}(\mathcal{X}_{\mathrm{en}}^{l-1})$, where $l \in\{1, \cdots, N\}$ denotes the output of $l$-th encoder layer and $\mathcal{X}_{\mathrm{en}}^{0} \in \mathbb{R}^{I \times D}$ is the embedded historical series. The $\text{Encoder}(\cdot)$ is formalized as
\begin{small}
\begin{equation}
\begin{aligned}
\mathcal{S}_{\mathrm{en}}^{l, 1},_{-}&=\text {\footnotesize MOEDecomp(\footnotesize FEB}\left(\mathcal{X}_{\mathrm{en}}^{l-1}\right)+\mathcal{X}_{\mathrm{en}}^{l-1}), \\
\mathcal{S}_{\mathrm{en}}^{l, 2},_{-}&=\text {\footnotesize MOEDecomp(\footnotesize FeedForward}\left(\mathcal{S}_{\mathrm{en}}^{l, 1}\right)+\mathcal{S}_{\mathrm{en}}^{l, 1}),\\
\mathcal{X}_{\text {en }}^{l}&=\mathcal{S}_{\text {en }}^{l, 2},
\end{aligned}
\end{equation}
\end{small}
where 
$\mathcal{S}_{\mathrm{en}}^{l, i}, i \in\{1,2\}$ represents the seasonal component after the $i$-th decomposition block in the $l$-th layer respectively. 
For FEB module, it has two different versions (FEB-f \& FEB-w) which are implemented through Discrete Fourier transform (DFT) and Discrete Wavelet transform (DWT) mechanism respectively and can seamlessly replace the self-attention block.

The decoder also adopts a multilayer structure as: $\mathcal{X}_{\mathrm{de}}^{l}, \mathcal{T}_{\mathrm{de}}^{l}=\text{Decoder}(\mathcal{X}_{\mathrm{de}}^{l-1}, \mathcal{T}_{\mathrm{de}}^{l-1})$, where $l \in\{1, \cdots, M\}$ denotes the output of $l$-th decoder layer. 
The $\text{Decoder}(\cdot)$ is formalized as
\begin{small}
\begin{equation}
\begin{aligned}
\mathcal{S}_{\mathrm{de}}^{l, 1}, \mathcal{T}_{\mathrm{de}}^{l, 1} &=\text{\footnotesize MOEDecomp}\left(\text {\footnotesize FEB}\left(\mathcal{X}_{\mathrm{de}}^{l-1}\right)+\mathcal{X}_{\mathrm{de}}^{l-1}\right), \\
\mathcal{S}_{\mathrm{de}}^{l, 2}, \mathcal{T}_{\mathrm{de}}^{l, 2} &=\text{\footnotesize MOEDecomp}\left(\text {\footnotesize FEA}\left(\mathcal{S}_{\mathrm{de}}^{l, 1}, \mathcal{X}_{\mathrm{en}}^{N}\right)+\mathcal{S}_{\mathrm{de}}^{l, 1}\right), \\
\mathcal{S}_{\mathrm{de}}^{l, 3}, \mathcal{T}_{\mathrm{de}}^{l, 3} &= \text{\footnotesize MOEDecomp}\left(\text {\footnotesize FeedForward}\left(\mathcal{S}_{\mathrm{de}}^{l, 2}\right)+\mathcal{S}_{\mathrm{de}}^{l, 2}\right), \\
\mathcal{X}_{\text {de}}^{l}  &= \mathcal{S}_{\text {de}}^{l, 3},\\
\mathcal{T}_{\mathrm{de}}^{l} &= \mathcal{T}_{\mathrm{de}}^{l-1} + \mathcal{W}_{l, 1} \!\cdot\! \mathcal{T}_{\mathrm{de}}^{l, 1}+\mathcal{W}_{l, 2} \!\cdot\! \mathcal{T}_{\mathrm{de}}^{l, 2}+\mathcal{W}_{l, 3} \!\cdot\! \mathcal{T}_{\mathrm{de}}^{l, 3},
\end{aligned}
\end{equation}
\end{small}
where $\mathcal{S}_{\text {de }}^{l, i}, \mathcal{T}_{\text {de }}^{l, i}, i \in\{1,2,3\}$ represent the seasonal and trend component after the $i$-th decomposition block in the $l$-th layer respectively. $\mathcal{W}_{l, i}, i \in\{1,2,3\}$ represents the projector for the $i$-th extracted trend $\mathcal{T}_{\text {de }}^{l, i}$.
Similar to FEB, FEA has two different versions (FEA-f \& FEA-w) which are implemented through DFT and DWT projection respectively with attention design, and can replace the cross-attention block.
The detailed description of $\text{FEA}(\cdot)$ will be given in the following Section~\ref{sec_Waveletblock}.

The final prediction is the sum of the two refined decomposed components as $\mathcal{W}_{\mathcal{S}} \cdot \mathcal{X}_{\mathrm{de}}^{M}+\mathcal{T}_{\mathrm{de}}^{M}$, where $\mathcal{W}_{\mathcal{S}}$ is to project the deep transformed seasonal component $\mathcal{X}_{\mathrm{de}}^{M}$ to the target dimension.

\subsection{Fourier Enhanced Structure}\label{sec_Fourierblock}


\paragraph{Discrete Fourier Transform (DFT)} The proposed Fourier Enhanced Structures use discrete Fourier transform (DFT). Let $\mathcal{F}$ denotes the Fourier transform and $\mathcal{F}^{-1}$ denotes the inverse Fourier transform. Given a sequence of real numbers $x_n$ in time domain, where $n=1,2...N$. DFT is defined as $X_l = \sum^{N-1}_{n=0} x_n e^{-i \omega ln}$,
where $i$ is the imaginary unit and $X_l$, $l=1,2...L$ is a sequence of complex numbers in the frequency domain. Similarly, the inverse DFT is defined as $x_n = \sum^{L-1}_{l=0} X_l e^{i \omega ln}$.
The complexity of DFT is $O(N^2)$. With fast Fourier transform (FFT), the computation complexity can be reduced to $O(N \log N)$. Here a random subset of the Fourier basis is used and the scale of the subset is bounded by a scalar. When we choose the mode index before DFT and reverse DFT operations, the computation complexity can be further reduced to $O(N)$.

\begin{figure}[t]
\centering
\includegraphics[width=\linewidth]{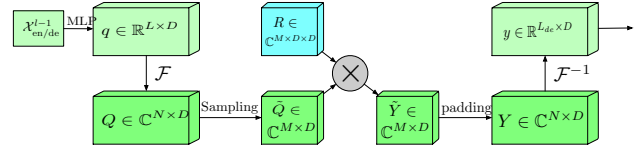}
\caption{Frequency Enhanced Block with Fourier transform (FEB-f) structure.}
\label{fig:FEB-f}
\end{figure}
\begin{figure}[t]
\centering
\includegraphics[width=\linewidth]{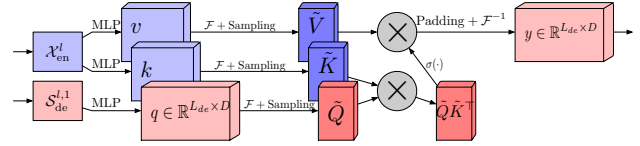}
\caption{Frequency Enhanced Attention with Fourier transform (FEA-f) structure, $\sigma(\cdot)$ is the activation function.}
\label{fig:FEA-f}
\vskip -0.1in
\end{figure}

\paragraph{Frequency Enhanced Block with Fourier Transform (FEB-f)}
The FEB-f is used in both encoder and decoder as shown in Figure \ref{fig:FEDformerStructure}. The input ($\bm{x} \in \mathbb{R}^{N \times D}$) of the FEB-f block is first linearly projected with $\bm{w} \in \mathbb{R}^{D \times D}$, so $\bm{q} = \bm{x} \cdot \bm{w}$. Then $\bm{q}$ is converted from the time domain to the frequency domain.
The Fourier transform of $\bm{q}$ is denoted as $\bm{Q} \in \mathbb{C}^{N \times D}$. In frequency domain, only the randomly selected $M$ modes are kept so we use a select operator as

\begin{small}
\begin{equation}
    \tilde{\bm{Q}}= {\rm \footnotesize Select} (\bm{Q}) = {\rm \footnotesize Select} (\mathcal{F}(\bm{q})),
\end{equation}
\end{small}

where $\tilde{\bm{Q}} \in \mathbb{C}^{M \times D}$ and $M << N$.
Then, the FEB-f is defined as
\begin{small}
\begin{equation}
    {\rm FEB\mbox{-}f} (\bm{q}) = 
    \mathcal{F}^{-1} ({\rm Padding} (\tilde{\bm{Q}} \odot \bm{R})),
\label{func:FNO-self}
\end{equation}
\end{small}
where 
$\bm{R} \in \mathbb{C}^{D \times D \times M}$ is a parameterized kernel initialized randomly. Let $\bm{Y} = \bm{Q} \odot \bm{C}$, with $\bm{Y} \in \mathbb{C}^{M \times D}$. The production operator $\odot$ is defined as: $Y_{m,d_o} = \sum _{d_i=0}^D Q_{m,d_i} \cdot R_{d_i,d_o,m}$, where $d_i=1,2...D$ is the input channel and $d_o=1,2...D$ is the output channel.
The result of $\bm{Q} \odot \bm{R}$ is then zero-padded to $\mathbb{C}^{N \times D}$ before performing inverse Fourier transform back to the time domain. The structure is shown in Figure \ref{fig:FEB-f}.

\paragraph{Frequency Enhanced Attention with Fourier Transform (FEA-f)}

We use the expression of the canonical transformer. The input: queries, keys, values are denoted as $\bm{q} \in \mathbb{R}^{L \times D}$, $\bm{k} \in \mathbb{R}^{L \times D}$, $\bm{v} \in \mathbb{R}^{L \times D}$. In cross-attention, the queries come from the decoder and can be obtained by $\bm{q} = \bm{x}_{en} \cdot \bm{w}_q$, where $\bm{w}_q \in \mathbb{R}^{D \times D}$. The keys and values are from the encoder and can be obtained by $\bm{k} = \bm{x}_{de} \cdot \bm{w}_k$ and $\bm{v} = \bm{x}_{de} \cdot \bm{w}_v$, where $\bm{w}_k, \bm{w}_v \in \mathbb{R}^{D \times D}$. Formally, the canonical attention can be written as
\begin{small}
\begin{equation}
    {\rm Atten}(\bm{q},\bm{k},\bm{v})=\text{Softmax}(\frac{\bm{q}\bm{k}^\top}{\sqrt{d_q}})\bm{v}.
\end{equation}
\end{small}
In FEA-f, we convert the queries, keys, and values with Fourier Transform and perform a similar attention mechanism in the frequency domain, by randomly selecting M modes. We denote the selected version after Fourier Transform as $\tilde{\bm{Q}} \in \mathbb{C}^{M \times D}$, $\tilde{\bm{K}} \in \mathbb{C}^{M \times D}$, $\tilde{\bm{V}} \in \mathbb{C}^{M \times D}$. The FEA-f is defined as
\begin{small} 
\begin{equation}
\begin{split}
{\tilde{\bm{Q}}}= {\rm Select} (\mathcal{F}(\bm{q}))\\ 
{\tilde{\bm{K}}}= {\rm Select} (\mathcal{F}(\bm{k}))\\ 
{\tilde{\bm{V}}}= {\rm Select} (\mathcal{F}(\bm{v}))
\end{split}
\end{equation}
\vskip -0.2in
\begin{equation}
    {\rm FEA\mbox{-}f} (\bm{q},\bm{k},\bm{v}) = \mathcal{F}^{-1}({\rm Padding}(\sigma( \tilde{\bm{Q}} \cdot {\tilde{\bm{K}}}^\top )\cdot \tilde{\bm{V}})),
\label{func:FNO-Corss}
\end{equation}
\end{small}
where $\sigma$ is the activation function. We use softmax or tanh for activation, since their converging performance differs in different data sets. Let $\bm{Y} = \sigma( \tilde{\bm{Q}} \cdot {\tilde{\bm{K}}}^\top )\cdot \tilde{\bm{V}}$, and $\bm{Y} \in \mathbb{C}^{M \times D}$ needs to be zero-padded to $\mathbb{C}^{L \times D}$ before performing inverse Fourier transform. The FEA-f structure is shown in Figure \ref{fig:FEA-f}.



\subsection{Wavelet Enhanced Structure}\label{sec_Waveletblock}
\paragraph{Discrete Wavelet Transform (DWT)}
While the Fourier transform creates a representation of the signal in the frequency domain, the Wavelet transform creates a representation in both the frequency and time domain, allowing efficient access of localized information of the signal. The multiwavelet transform synergizes the advantages of orthogonal polynomials as well as wavelets. 
For a given $f(x)$, the multiwavelet coefficients at the scale $n$ can be defined as $\mathbf{s}_{l}^{n}=\left[\left\langle f, \phi_{i l}^{n}\right\rangle_{\mu_{n}}\right]_{i=0}^{k-1}$, $\mathbf{d}_{l}^{n}=\left[\left\langle f, \psi_{i l}^{n}\right\rangle_{\mu_{n}}\right]_{i=0}^{k-1}$, respectively, w.r.t. measure $\mu_{n}$ with $\mathbf{s}_{l}^{n}, \mathbf{d}_{l}^{n} \in \mathbb{R}^{k \times 2^{n}}$. $\phi_{i l}^{n}$ are wavelet orthonormal basis of piecewise polynomials. 
The decomposition/reconstruction across scales is defined as

\begin{small} 
\begin{equation}
\begin{aligned}
\mathbf{s}_{l}^{n} &=H^{(0)} \mathbf{s}_{2 l}^{n+1}+H^{(1)} \mathbf{s}_{2 l+1}^{n+1}, \\
\mathbf{s}_{2 l}^{n+1} &=\Sigma^{(0)}\left(H^{(0) T} \mathbf{s}_{l}^{n}+G^{(0) T} \mathbf{d}_{l}^{n}\right), \\
\mathbf{d}_{l}^{n} &=G^{(0)} \mathbf{s}_{2 l}^{n+1}+H^{(1)} \mathbf{s}_{2 l+1}^{n+1}, \\
\mathbf{s}_{2 l+1}^{n+1} &=\Sigma^{(1)}\left(H^{(1) T} \mathbf{s}_{l}^{n}+G^{(1) T} \mathbf{d}_{l}^{n}\right), 
\end{aligned}
\end{equation}
\end{small}

where $\left(H^{(0)}, H^{(1)}, G^{(0)}, G^{(1)}\right)$ are linear coefficients for multiwavelet decomposition filters. They are fixed matrices used for wavelet decomposition. 
The multiwavelet representation of a signal can be obtained by the tensor product of multiscale and multiwavelet basis. Note that the basis at various scales are coupled by the tensor product, so we need to untangle it. Inspired by \cite{Multiwavelet-based-Operator-Learning}, we adapt a non-standard wavelet representation to reduce the model complexity. For a map function $F(x)=x'$, the map under multiwavelet domain can be written as
\begin{small}
\begin{equation}
U_{d l}^{n}=A_{n} d_{l}^{n}+B_{n} s_{l}^{n}, \quad U_{\hat{s} l}^{n}=C_{n} d_{l}^{n}, \quad U_{s l}^{L}=\bar{F}s_{l}^{L},
\end{equation}
\end{small}
where $\left(U_{s l}^{n}, U_{d l}^{n}, s_{l}^{n}, d_{l}^{n}\right)$ are the multiscale, multiwavelet coefficients, $L$ is the coarsest scale under recursive decomposition, and $A_{n}$, $B_{n}$, $C_{n}$ are three independent FEB-f blocks modules used for processing different signal during decomposition and reconstruction. Here $\bar{F}$ is a single-layer of perceptrons which processes the remaining coarsest signal after $L$ decomposed steps.
More designed detail is described in Appendix \ref{app:wavelets}.
\begin{figure}[t]
\centering
\setlength{\abovecaptionskip}{-5pt}
\setlength{\belowcaptionskip}{-10pt} 
\scalebox{0.95}{
\includegraphics[width=\linewidth]{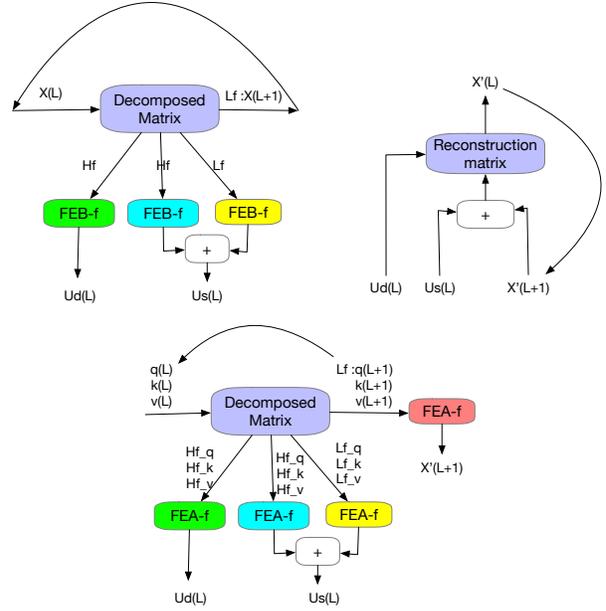}}
\caption{Top Left: Wavelet frequency enhanced block decomposition stage. Top Right: Wavelet block reconstruction stage shared by FEB-w and FEA-w. Bottom: Wavelet frequency enhanced cross attention decomposition stage.}
\label{fig:feb-w}
\vskip -0.1in
\end{figure}

\paragraph{Frequency Enhanced Block with Wavelet Transform (FEB-w)}
The overall FEB-w architecture is shown in Figure \ref{fig:feb-w}. It differs from FEB-f in the recursive mechanism: the input is decomposed into 3 parts recursively and operates individually. For the wavelet decomposition part, we implement the fixed Legendre wavelets basis decomposition matrix. Three FEB-f modules are used to process the resulting high-frequency part, low-frequency part, and remaining part from wavelet decomposition respectively. For each cycle $L$, it produces a processed high-frequency tensor $Ud(L)$, a processed low-frequency frequency tensor $Us(L)$, and the raw low-frequency tensor $X(L+1)$. This is a ladder-down approach, and the decomposition stage performs the decimation of the signal by a factor of 1/2, running for a maximum of $L$ cycles, where $L<\log _{2}(M)$ for a given input sequence of size $M$. In practice, $L$ is set as a fixed argument parameter. The three sets of FEB-f blocks are shared during different decomposition cycles $L$. 
For the wavelet reconstruction part, we recursively build up our output tensor as well. For each cycle $L$, we combine $X(L+1)$, $Us(L)$, and $Ud(L)$ produced from the decomposition part and produce $X(L)$ for the next reconstruction cycle. For each cycle, the length dimension of the signal tensor is increased by 2 times. 
\paragraph{Frequency Enhanced Attention with Wavelet Transform (FEA-w)}
FEA-w contains the decomposition stage and reconstruction stage like FEB-w. Here we keep the reconstruction stage unchanged. The only difference lies in the decomposition stage. The same decomposed matrix is used to decompose $\bm{q},\bm{k},\bm{v}$ signal separately, and $\bm{q},\bm{k},\bm{v}$ share the same sets of module to process them as well. As shown above, a frequency enhanced block with wavelet decomposition block (FEB-w) contains three FEB-f blocks for the signal process. We can view the FEB-f as a substitution of self-attention mechanism. We use a straightforward way to build the frequency enhanced cross attention with wavelet decomposition, substituting each FEB-f with a FEA-f module. Besides, another FEA-f module is added to process the coarsest remaining $q(L),k(L),v(L)$ signal.

\subsection{Mixture of Experts for Seasonal-Trend Decomposition}\label{sec_MOEDecomp}
Because of the commonly observed complex periodic pattern coupled with the trend component on real-world data, extracting the trend can be hard with fixed window average pooling. To overcome such a problem, we design a Mixture Of Experts Decomposition block (MOEDecomp). It contains a set of average filters with different sizes to extract multiple trend components from the input signal and a set of data-dependent weights for combining them as the final trend. Formally, we have

\begin{small}
\begin{equation}
    \mathbf{X_{trend}} = \mathbf{Softmax}(L(x))*(F(x)),
\end{equation}
\end{small}
where $F(\cdot)$ is a set of average pooling filters and Softmax($L(x)$) is the weights for mixing these extracted trends.

\subsection{Complexity Analysis}
For FEDformer-f, the computational complexity for time and memory is $O(L)$ with a fixed number of randomly selected modes in FEB \& FEA blocks. We set modes number $M=64$ as default value. 
Though the complexity of full DFT transformation by FFT is ($O(L\log (L)$), 
our model only needs $O(L)$ cost and memory complexity with the pre-selected set of Fourier basis for quick implementation.
For FEDformer-w, when we set the recursive decompose step to a fixed number $L$ and use a fixed number of randomly selected modes the same as FEDformer-f, the time complexity and memory usage are $O(L)$ as well. In practice, we choose $L=3$ and modes number $M=64$ as default value. 
The comparisons of the time complexity and memory usage in training and the inference steps in testing are summarized in Table \ref{tab:compare_complexity}. It can be seen that the proposed FEDformer achieves the best overall complexity among Transformer-based forecasting models.

\begin{table}
\centering
\caption{Complexity analysis of different forecasting models.} 
\label{tab:compare_complexity}
\vskip 0.05in
\scalebox{0.8}{
\begin{tabular}{l|c|c|c}
\hline \multirow{2}{*}{ Methods } & \multicolumn{2}{|c}{ Training } & Testing \\
\cline { 2 - 4 } & Time & Memory & Steps \\
\hline FEDformer & $\mathcal{O}(L)$ & $\mathcal{O}(L)$ & 1 \\
\hline Autoformer & $\mathcal{O}(L \log L)$ & $\mathcal{O}(L \log L)$ & 1 \\
\hline Informer & $\mathcal{O}(L \log L)$ & $\mathcal{O}(L \log L)$ & 1 \\
\hline Transformer & $\mathcal{O}\left(L^{2}\right)$ & $\mathcal{O}\left(L^{2}\right)$ & $L$ \\
\hline LogTrans & $\mathcal{O}(L \log L)$ & $\mathcal{O}\left(L^{2}\right)$ & 1 \\
\hline Reformer & $\mathcal{O}(L \log L)$ & $\mathcal{O}(L \log L)$ & $L$ \\
\hline LSTM & $\mathcal{O}(L)$ & $\mathcal{O}(L)$ & $L$ \\
\hline
\end{tabular}
}
\vskip -0.1in
\end{table}




\section{Experiments}
To evaluate the proposed FEDformer, we conduct extensive experiments on six popular real-world datasets, including energy, economics, traffic, weather, and disease. 
Since classic models like ARIMA and basic RNN/CNN models perform relatively inferior as shown in \cite{haoyietal-informer-2021} and~\cite{Autoformer}, we mainly include four state-of-the-art transformer-based models for comparison, i.e., Autoformer~\cite{Autoformer}, Informer~\cite{haoyietal-informer-2021}, LogTrans~\cite{Log-transformer-shiyang-2019} and Reformer~\cite{DBLP:conf/iclr/KitaevKL20-reformer} as baseline models. 
Note that since Autoformer holds the best performance in all the six benchmarks, it is used as the main baseline model for comparison.  
More details about baseline models, datasets, and implementation are described in Appendix \ref{App:related_work_transformers}, \ref{app:exp:dataset}, and \ref{app:exp:implement}, respectively.

\subsection{Main Results}
For better comparison, we follow the experiment settings of Autoformer in \cite{Autoformer} where the input length is fixed to 96, and the prediction lengths for both training and evaluation are fixed to be 96, 192, 336, and 720, respectively. 




\begin{table*}[t]
\setlength\tabcolsep{3pt} 
\centering
\begin{small}
\caption{Multivariate long-term series forecasting results on six datasets with input length $I=96$ and prediction length $O \in \{96,192,336,720\}$ (For ILI dataset, we use input length $I=36$ and prediction length $O \in \{24,36,48,60\}$). A lower MSE indicates better performance, and the best results are highlighted in bold.}
\vskip 0.05in
\scalebox{0.65}{
\begin{tabular}{c|c|cccc|cccc|cccc|cccc|cccc|cccc}
\toprule
\multirow{2}{*}{Methods} & \multirow{2}{*}{Metric} &\multicolumn{4}{c|}{ETTm2}&\multicolumn{4}{c|}{Electricity}&\multicolumn{4}{c|}{Exchange}&\multicolumn{4}{c|}{Traffic}&\multicolumn{4}{c|}{Weather}&\multicolumn{4}{c}{ILI}\\
& &96&192&336&720&96& 192 & 336 & 720 & 96 & 192 & 336 & 720
& 96 & 192 & 336 & 720
& 96 & 192 & 336 & 720& 24 & 36 & 48 & 60\\
\midrule
\multirow{2}{*}{FEDformer-f} & MSE&\textbf{0.203}&\textbf{0.269} &\textbf{0.325} &\textbf{0.421} &0.193 &0.201 &0.214 &0.246 &0.148 &0.271 &0.460 &1.195 &0.587& 0.604& 0.621& 0.626 &\textbf{0.217} &\textbf{0.276} &\textbf{0.339} &\textbf{0.403} &3.228 &2.679 &2.622 &2.857\\
& MAE &\textbf{0.287}&\textbf{0.328}&\textbf{0.366}&\textbf{0.415}&0.308&0.315&0.329&0.355&0.278&0.380&0.500&0.841&0.366&0.373&0.383&0.382&\textbf{0.296}&\textbf{0.336}&\textbf{0.380}&\textbf{0.428}&1.260&1.080&1.078&1.157 \\\midrule

\multirow{2}{*}{FEDformer-w} & MSE&0.204&0.316&0.359&0.433&\textbf{0.183}&\textbf{0.195}&\textbf{0.212}&\textbf{0.231}&\textbf{0.139}&\textbf{0.256}&\textbf{0.426}&\textbf{1.090}&\textbf{0.562}&\textbf{0.562}&\textbf{0.570}&\textbf{0.596}&0.227&0.295&0.381&0.424&\textbf{2.203}&\textbf{2.272}&\textbf{2.209}&\textbf{2.545}\\

& MAE &0.288&0.363&0.387&0.432& \textbf{0.297}& \textbf{0.308} &\textbf{0.313} &\textbf{0.343} &\textbf{0.276}&\textbf{0.369}&\textbf{0.464}&\textbf{0.800}&\textbf{0.349}&\textbf{0.346}&\textbf{0.323}&\textbf{0.368}&0.304&0.363&0.416&0.434&\textbf{0.963}&\textbf{0.976}&\textbf{0.981}&\textbf{1.061} \\\midrule
\multirow{2}{*}{Autoformer} & MSE&0.255&0.281&0.339&0.422&0.201&0.222&0.231&0.254&0.197&0.300&0.509&1.447&0.613&0.616&0.622&0.660&0.266&0.307&0.359&0.419&3.483&3.103&2.669&2.770\\
& MAE &0.339&0.340&0.372&0.419&0.317&0.334&0.338&0.361&0.323&0.369&0.524&0.941&0.388&0.382&0.337&0.408&0.336&0.367&0.395&0.428&1.287&1.148&1.085&1.125 \\\midrule
\multirow{2}{*}{Informer} & MSE&0.365&0.533&1.363&3.379&0.274&0.296&0.300&0.373&0.847&1.204&1.672&2.478&0.719&0.696&0.777&0.864&0.300&0.598&0.578&1.059&5.764&4.755&4.763&5.264\\
& MAE &0.453&0.563&0.887&1.338&0.368&0.386&0.394&0.439&0.752&0.895&1.036&1.310&0.391&0.379&0.420&0.472&0.384&0.544&0.523&0.741&1.677&1.467&1.469&1.564 \\\midrule

\multirow{2}{*}{LogTrans} & MSE&0.768&0.989&1.334&3.048 &0.258& 0.266& 0.280& 0.283& 0.968& 1.040& 1.659& 1.941& 0.684& 0.685& 0.7337& 0.717& 0.458& 0.658& 0.797&0.869& 4.480& 4.799& 4.800& 5.278\\
& MAE &0.642&0.757&0.872 & 1.328& 0.357 &0.368 &0.380& 0.376 & 0.812 &0.851& 1.081& 1.127& 0.384& 0.390& 0.408& 0.396& 0.490 &0.589& 0.652& 0.675& 1.444& 1.467& 1.468& 1.560\\\midrule
\multirow{2}{*}{Reformer} & MSE&0.658& 1.078 & 1.549 & 2.631 & 0.312 & 0.348 & 0.350 & 0.340 & 1.065 &1.188 & 1.357 & 1.510 & 0.732 & 0.733 &0.742 & 0.755 & 0.689 & 0.752 & 0.639 & 1.130 & 4.400 & 4.783 & 4.832 & 4.882\\
& MAE &0.619&0.827 &0.972 & 1.242 & 0.402 & 0.433 & 0.433 & 0.420 & 0.829 & 0.906 & 0.976 & 1.016 & 0.423 & 0.420 & 0.420 & 423 & 0.596 & 0.638 & 0.596 & 0.792 & 1.382 & 1.448 & 1.465 & 1.483 \\

\bottomrule
\end{tabular}
\label{tab:multi-benchmarks}
}
\end{small}
\vskip -0.1in
\end{table*}

\begin{table*}[t]
\setlength\tabcolsep{3pt} 
\centering
\begin{small}
\caption{Univariate long-term series forecasting results on six datasets with input length $I=96$ and prediction length $O \in \{96,192,336,720\}$ (For ILI dataset, we use input length $I=36$ and prediction length $O \in \{24,36,48,60\}$). A lower MSE indicates better performance, and the best results are highlighted in bold.}
\vskip 0.05in
\scalebox{0.65}{
\begin{tabular}{c|c|cccc|cccc|cccc|cccc|cccc|cccc}
\toprule
\multirow{2}{*}{Methods} & \multirow{2}{*}{Metric} &\multicolumn{4}{c|}{ETTm2}&\multicolumn{4}{c|}{Electricity}&\multicolumn{4}{c|}{Exchange}&\multicolumn{4}{c|}{Traffic}&\multicolumn{4}{c|}{Weather}&\multicolumn{4}{c}{ILI}\\
& & 96 & 192 & 336 & 720 & 96 & 192 & 336 & 720 & 96 & 192 & 336 & 720
& 96 & 192 & 336 & 720
& 96 & 192 & 336 & 720 & 24 & 36 & 48 & 60\\
\midrule
\multirow{2}{*}{FEDformer-f} & MSE & 0.072 &  \textbf{0.102} &  \textbf{0.130} &  \textbf{0.178} &  \textbf{0.253} &  \textbf{0.282} &  \textbf{0.346} &  \textbf{0.422} & 0.154 &  0.286 &  0.511 &  1.301 &   0.207 &  0.205 &  0.219 &  0.244 &   0.0062 &  0.0060 &  \textbf{0.0041} &  \textbf{0.0055} &  0.708 &  0.584 &  0.717 &  0.855 \\
& MAE &  0.206 &  \textbf{0.245} &  \textbf{0.279} &  \textbf{0.325} &  \textbf{0.370} &  \textbf{0.386} &  \textbf{0.431} &  \textbf{0.484} & 0.304 &  0.420 &  0.555 &  0.879 & 0.312 &  0.312 &  0.323 &  0.344 &   0.062 &  0.062 &  \textbf{0.050} &  \textbf{0.059} &  0.627 &  0.617 &  0.697 &  0.774\\\midrule

\multirow{2}{*}{FEDformer-w} & MSE&  \textbf{0.063} & 0.110 & 0.147 & 0.219 & 0.262 & 0.316 & 0.361 & 0.448 & \textbf{0.131} & \textbf{0.277} & \textbf{0.426} & \textbf{1.162} & \textbf{0.170} & \textbf{0.173} & \textbf{0.178} & \textbf{0.187} & \textbf{0.0035} & \textbf{0.0054} & 0.008 & 0.015 & \textbf{0.693} & \textbf{0.554} & \textbf{0.699} & \textbf{0.828}\\

& MAE &\textbf{0.189} & 0.252 & 0.301 & 0.368 & 0.378 & 0.410 & 0.445 & 0.501 & \textbf{0.284} &\textbf{ 0.420} & \textbf{0.511} & \textbf{0.832} & \textbf{0.263} & \textbf{0.265} & \textbf{0.266} & \textbf{0.286} & \textbf{0.046} & \textbf{0.059} & 0.072 & 0.091 & \textbf{0.629} & \textbf{0.604} & \textbf{0.696} & \textbf{0.770}\\\midrule
\multirow{2}{*}{Autoformer} & MSE & 0.065 & 0.118 & 0.154 & 0.182 & 0.341 & 0.345 & 0.406 & 0.565 & 0.241 & 0.300 & 0.509 & 1.260 & 0.246 & 0.266 & 0.263 & 0.269 & 0.011 & 0.0075 & 0.0063 & 0.0085 & 0.948 & 0.634 & 0.791 & 0.874 \\
& MAE & 0.189 & 0.256 & 0.305 & 0.335 & 0.438 & 0.428 & 0.470 & 0.581 & 0.387 & 0.369 & 0.524 & 0.867 & 0.346 & 0.370 & 0.371 & 0.372 & 0.081 & 0.067 & 0.062 & 0.070 & 0.732 & 0.650 & 0.752 & 0.797\\\midrule
\multirow{2}{*}{Informer} & MSE & 0.080 &  0.112 &  0.166 &  0.228 & 0.258 & 0.285 & 0.336 & 0.607 & 1.327 & 1.258 & 2.179 & 1.280 & 0.257 & 0.299 & 0.312 & 0.366 & 0.004 & 0.002 & 0.004 & 0.003 & 5.282 & 4.554 & 4.273 & 5.214\\
& MAE & 0.217 &  0.259 &  0.314 &  0.380 & 0.367 & 0.388 & 0.423 & 0.599 & 0.944 & 0.924 & 1.296 & 0.953 & 0.353 & 0.376 & 0.387 & 0.436 & 0.044 & 0.040 & 0.049 & 0.042 & 2.050 & 1.916 & 1.846 & 2.057\\\midrule
\multirow{2}{*}{LogTrans} & MSE & 0.075& 0.129&	0.154&	0.160&	0.288&	0.432&	0.430&	0.491&	0.237&	0.738&	2.018&	2.405&	0.226&	0.314&	0.387&	0.437&	0.0046& 0.0060&	0.0060&	0.007&	3.607&	2.407&	3.106&	3.698 \\
& MAE & 0.208&	0.275&	0.302&	0.322&	0.393&	0.483&	0.483&	0.531&	0.377&	0.619&	1.070&	1.175&	0.317&	0.408&	0.453&	0.491&	0.052&	0.060&	0.054&	0.059&	1.662&	1.363&	1.575&	1.733\\\midrule
\multirow{2}{*}{Reformer} & MSE &  0.077 &  0.138 &  0.160 &  0.168 &  0.275 &  0.304 &  0.370 &  0.460 &    0.298 &  0.777 &  1.833 &  1.203 &   0.313 &  0.386 &  0.423 &  0.378 &   0.012 &  0.0098 &  0.013 &  0.011 &  3.838 &  2.934 &  3.755 &  4.162\\

& MAE &  0.214 &  0.290 &  0.313 &  0.334 &  0.379 &  0.402 &  0.448 &  0.511 &    0.444 &  0.719 &  1.128 &  0.956 &   0.383 &  0.453 &  0.468 &  0.433 &   0.087 &  0.044 &  0.100 &  0.083 &  1.720 &  1.520 &  1.749 &  1.847\\
\bottomrule

\end{tabular}
}
\label{tab:uni-benchmarks-large}
\end{small}
\vskip -0.1in
\end{table*}

\paragraph{Multivariate Results}
For the multivariate forecasting, FEDformer achieves the best performance on all six benchmark datasets at all horizons as shown in Table \ref{tab:multi-benchmarks}. Compared with Autoformer, the proposed FEDformer yields an overall \textbf{14.8\%} relative MSE reduction. It is worth noting that for some of the datasets, such as Exchange and ILI, the improvement is even more significant (over $20\%$). Note that the Exchange dataset does not exhibit clear periodicity in its time series, but FEDformer can still achieve superior performance. 
Overall, the improvement made by FEDformer is consistent with varying horizons, implying its strength in long term forecasting.  
More detailed results on ETT full benchmark are provided in Appendix \ref{app:exp:ett_benchmark}. 
\paragraph{Univariate Results}


The results for univariate time series forecasting are summarized in Table \ref{tab:uni-benchmarks-large}. 
Compared with Autoformer, FEDformer yields an overall \textbf{22.6\%} relative MSE reduction, and on some datasets, such as traffic and weather, the improvement can be more than $30\%$. It again verifies that FEDformer is more effective in long-term forecasting. Note that due to the difference between Fourier and wavelet basis, FEDformer-f and FEDformer-w perform well on different datasets, making them complementary choice for long term forecasting. 
More detailed results on ETT full benchmark are provided in Appendix \ref{app:exp:ett_benchmark}. 

%


%




\subsection{Ablation Studies}
In this section, the ablation experiments are conducted, aiming at comparing the performance of frequency enhanced block and its alternatives. 
The current SOTA results of Autoformer which uses the autocorrelation mechanism serve as the baseline. Three ablation variants of FEDformer are tested: 1) FEDformer V1: we use FEB to substitute self-attention only; 2) FEDformer V2: we use FEA to substitute cross attention only; 3) FEDFormer V3: we use FEA to substitute both self and cross attention.
The ablated versions of FEDformer-f as well as the SOTA models are compared in Table \ref{tab:ablation}, and we use a bold number if the ablated version brings improvements compared with Autoformer. We omit the similar results in FEDformer-w due to space limit.
It can be seen in Table \ref{tab:ablation} that FEDformer V1 brings improvement in 10/16 cases, while FEDformer V2 improves in 12/16 cases. The best performance is achieved in our FEDformer with FEB and FEA blocks which improves performance in all 16/16 cases. This verifies the effectiveness of the designed FEB, FEA for substituting self and cross attention. 
Furthermore, experiments on ETT and Weather datasets show that the adopted MOEDecomp (mixture of experts decomposition) scheme can bring an average of 2.96\% improvement compared with the single decomposition scheme. More details are provided in Appendix \ref{app:moe_ablation}.



\begin{table*}[t]
\centering
\caption{Ablation studies: multivariate long-term series forecasting results on ETTm1 and ETTm2 with input length $I=96$ and prediction length $O \in \{96,192,336,720\}$. Three variants of FEDformer-f are compared with baselines. The best results are highlighted in bold.}
\label{sample-table}
\begin{center}
\begin{small}
\scalebox{0.80}{
\begin{tabular}{c|c|cccccccccccccc}
\toprule
\multicolumn{2}{c|}{Methods}&\multicolumn{2}{c|}{Transformer}&\multicolumn{2}{c|}{Informer}&\multicolumn{2}{c|}{Autoformer}&\multicolumn{2}{c|}{FEDformer V1}&\multicolumn{2}{c|}{FEDformer V2}&\multicolumn{2}{c|}{FEDformer V3}&\multicolumn{2}{c}{FEDformer-f }\\
\midrule
\multicolumn{2}{c|}{Self-att} &\multicolumn{2}{c|}{FullAtt} &\multicolumn{2}{c|}{ProbAtt} & \multicolumn{2}{c|}{AutoCorr} & \multicolumn{2}{c|}{FEB-f(Eq. \ref{func:FNO-self})} & \multicolumn{2}{c|}{AutoCorr}& \multicolumn{2}{c|}{FEA-f(Eq. \ref{func:FNO-Corss})} & \multicolumn{2}{c}{FEB-f(Eq. \ref{func:FNO-self})}\\

\multicolumn{2}{c|}{Cross-att} &\multicolumn{2}{c|}{FullAtt} &\multicolumn{2}{c|}{ProbAtt} & \multicolumn{2}{c|}{AutoCorr} & \multicolumn{2}{c|}{AutoCorr} & \multicolumn{2}{c|}{FEA-f(Eq. \ref{func:FNO-Corss})} & \multicolumn{2}{c|}{FEA-f(Eq. \ref{func:FNO-Corss})}& \multicolumn{2}{c}{FEA-f(Eq. \ref{func:FNO-Corss})}\\
\midrule

\multicolumn{2}{c|}{Metric} & MSE  & MAE & MSE & MAE& MSE  & MAE & MSE  & MAE & MSE  & MAE & MSE  & MAE & MSE  & MAE\\
\midrule
\multirow{4}{*}{\rotatebox{90}{ETTm1}} 
& 96 & 0.525 & 0.486 & 0.458 & 0.465 & 0.481 & 0.463 & \textbf{0.378} & \textbf{0.419} &  0.539 & 0.490 & 0.534 & 0.482 & \textbf{0.379} & \textbf{0.419}\\

& 192 & 0.526 & 0.502 & 0.564 & 0.521 & 0.628 & 0.526 & \textbf{0.417} & \textbf{0.442} & \textbf{0.556} & \textbf{0.499} & \textbf{0.552} & \textbf{0.493} & \textbf{0.426} & \textbf{0.441}\\
& 336 & 0.514 & 0.502 & 0.672 & 0.559 & 0.728 & 0.567 & \textbf{0.480} & \textbf{0.477} & \textbf{0.541} & \textbf{0.498} & \textbf{0.565} & \textbf{0.503} & \textbf{0.445} & \textbf{0.459}\\
& 720 & 0.564 & 0.529 & 0.714 & 0.596  & 0.658 & 0.548 & \textbf{0.543} & \textbf{0.517} & \textbf{0.558} & \textbf{0.507} & \textbf{0.585} & \textbf{0.515} & \textbf{0.543} & \textbf{0.490}\\
\midrule
\multirow{4}{*}{\rotatebox{90}{ETTm2}} 

& 96 & 0.268 & 0.346 & 0.227 & 0.305 & 0.255 & 0.339 & 0.259  & \textbf{0.337}& \textbf{0.216} &\textbf{0.297} & \textbf{0.211} & \textbf{0.292} &\textbf{0.203} & \textbf{0.287}      \\

& 192 & 0.304 & 0.355 & 0.300 & 0.360 & 0.281 & 0.340 & 0.285 & 0.344& \textbf{0.274} &\textbf{0.331} & \textbf{0.272} & \textbf{0.329} & \textbf{0.269} & \textbf{0.328}      \\

& 336 & 0.365 & 0.400 & 0.382 & 0.410 & 0.339 & 0.372 & \textbf{0.320} & 0.373& \textbf{0.334} &\textbf{0.369} & \textbf{0.327} & \textbf{0.363} & \textbf{0.325} & \textbf{0.366}\\

& 720 & 0.475 & 0.466 & 1.637 & 0.794 & 0.422 & 0.419 & 0.761 & 0.628& 0.427 &0.420 & \textbf{0.418} & \textbf{0.415} & \textbf{0.421} & \textbf{0.415} \\
\midrule
\multicolumn{2}{c|}{Count}& 0 & 0 & 0 & 0 & 0 & 0 & 5 & 5 & 6 & 6 & 7 & 7 & 8 & 8\\
\bottomrule
\end{tabular}
\label{tab:ablation}
}
\end{small}
\end{center}
\vskip -0.1in
\end{table*}

\subsection{Mode Selection Policy}

\begin{figure}[t]
\setlength{\abovecaptionskip}{-5pt}
\setlength{\belowcaptionskip}{-10pt} 
\begin{minipage}{\linewidth/2}
    \centering
    \setlength{\abovecaptionskip}{-5pt}
    \setlength{\belowcaptionskip}{-10pt}
    \includegraphics[width=\linewidth]{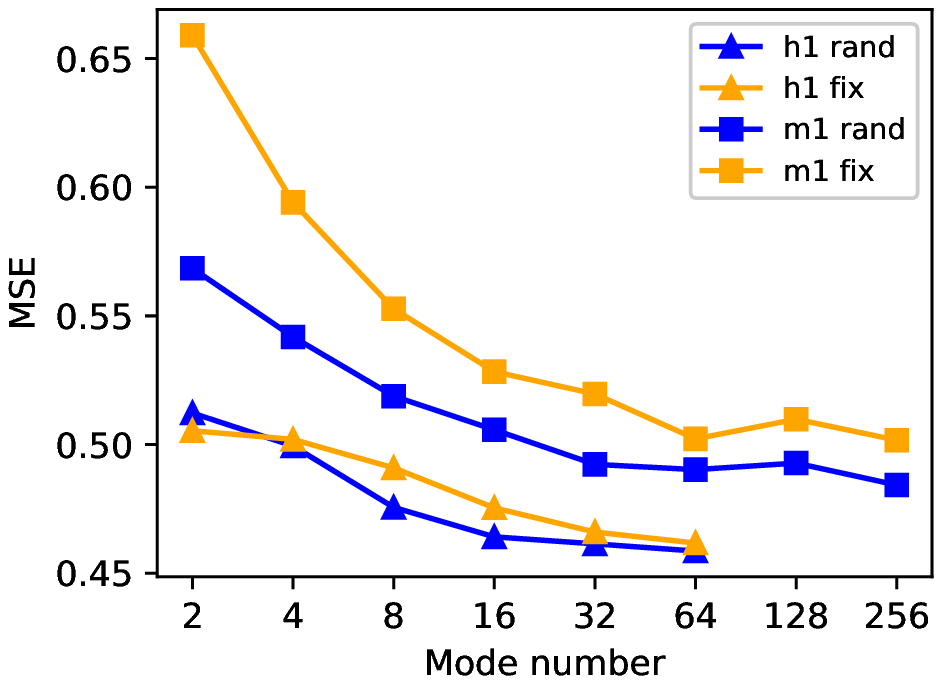}
    \end{minipage}\hfill
    \begin{minipage}{\linewidth/2}
    \centering
    \setlength{\abovecaptionskip}{-5pt}
    \setlength{\belowcaptionskip}{-10pt}
    \includegraphics[width=\linewidth]{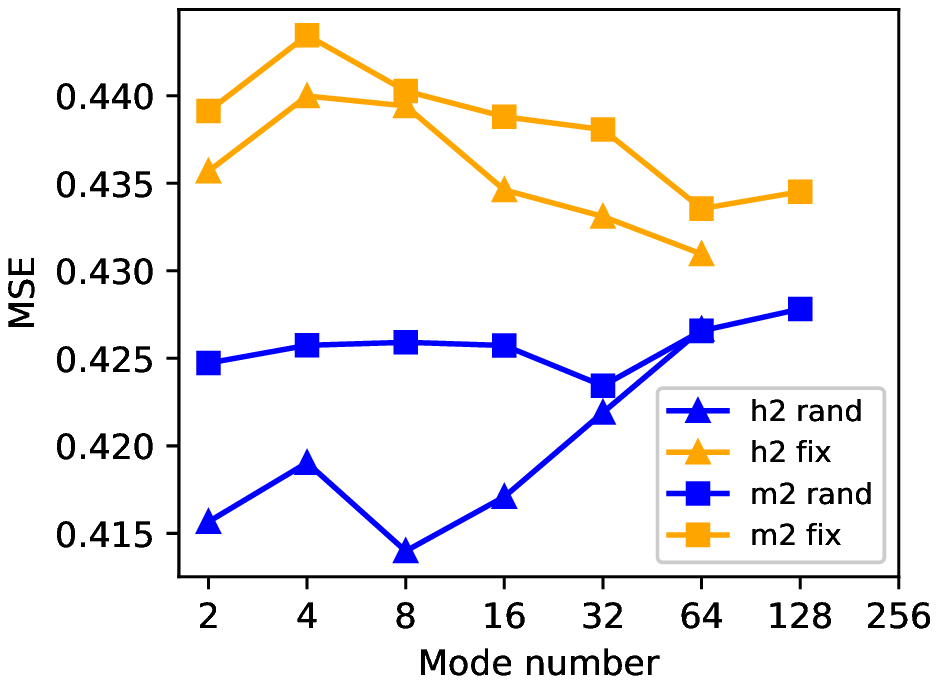}
    \end{minipage}\hfill
    \caption{Comparison of two base-modes selection method (Fix\&Rand). Rand policy means randomly selecting a subset of modes, Fix policy means selecting the lowest frequency modes. Two policies are compared on a variety of base-modes number $M \in \{2,4,8...256\}$ on ETT full-benchmark (h1, m1, h2, m2).}
    \label{fig_model_h1m1_h2m2_2}
    \vskip -0.2in
\end{figure}

The selection of discrete Fourier basis is the key to effectively representing the signal and maintaining the model's \textbf{linear} complexity.
As we discussed in Section \ref{sec:signal_represent_random_f_comp}, random Fourier mode selection is a better policy in forecasting tasks. more importantly, random policy requires no prior knowledge of the input and generalizes easily in new tasks. Here we empirically compare the random selection policy with fixed selection policy, and summarize the experimental results in Figure \ref{fig_model_h1m1_h2m2_2}.
It can be observed that the adopted random policy achieves better performance than the common fixed policy which only keeps the low frequency modes. Meanwhile, the random policy exhibits some \textbf{mode saturation effect}, indicating an appropriate random number of modes instead of all modes would bring better performance, which is also consistent with the theoretical analysis in Section~\ref{sec:signal_represent_random_f_comp}.

%


\subsection{Distribution Analysis of Forecasting Output}
In this section, we evaluate the distribution similarity between the input sequence and forecasting output of different transformer models quantitatively.
In Table \ref{tab:KStest_small}, we applied the Kolmogrov-Smirnov test to check if the forecasting results of different models made on ETTm1 and ETTm2 are consistent with the input sequences. In particular, we test if the
input sequence of fixed 96-time steps come from the same distribution as the predicted sequence, with the null hypothesis that both sequences come from the same distribution. 
On both datasets, by setting the common P-value as 0.01, various existing Transformer baseline models have much less values than 0.01 except Autoformer, which indicates their forecasting output have a higher probability to be sampled from the different distributions compared to the input sequence. In contrast, Autoformer and FEDformer have much larger P-value compared to others, which mainly contributes to their seasonal-trend decomposition mechanism. Though we get close results from ETTm2 by both models, the proposed FEDformer has much larger P-value in ETTm1. And it's the only model whose null hypothesis can not be rejected with P-value larger than 0.01 in all cases of the two datasets, implying that the output sequence generated by FEDformer shares a more similar distribution as the input sequence than others and thus justifies the our design motivation of FEDformer as discussed in Section~\ref{sec_intro}.
More detailed analysis are provided in Appendix \ref{app:output_dis_analy}.

\begin{table}[t]
\label{sample-table-kstest}
\centering
\caption{P-values of Kolmogrov-Smirnov test of different transformer models for long-term forecasting output on ETTm1 and ETTm2 dataset. Larger value indicates the hypothesis (the input sequence and forecasting output come from the same distribution) is less likely to be rejected. The best results are highlighted.}
\begin{small}
\vskip 0.05in
\scalebox{0.80}{
\begin{tabular}{c|c|ccccc}
\toprule
\multicolumn{2}{c|}{Methods}&Transformer&Informer&Autoformer&FEDformer&True\\ 
\midrule
\multirow{4}{*}{\rotatebox{90}{ETTm1}} 
& 96 & 0.0090 & 0.0055&0.020 & \textbf{0.048} & {0.023} \\
& 192 & 0.0052 & 0.0029&0.015 & \textbf{0.028} & {0.013} \\
& 336 & 0.0022 & 0.0019&0.012 & \textbf{0.015} & {0.010} \\
& 720 & 0.0023 & 0.0016&0.008 & \textbf{0.014} & {0.004} \\
\midrule
\multirow{4}{*}{\rotatebox{90}{ETTm2}} 
& 96 & 0.0012 & 0.0008& \textbf{0.079} & {0.071} & {0.087} \\
& 192 & 0.0011 & 0.0006& \textbf{0.047} & {0.045} & {0.060} \\
& 336 & 0.0005 & 0.00009& 0.027 & \textbf{0.028} & {0.042} \\
& 720 & 0.0008& 0.0002& \textbf{0.023} & {0.021} & {0.023} \\\midrule
\multicolumn{2}{c|}{Count}& 0 & 0 & 3 & 5 & NA \\
\bottomrule
\end{tabular}
}
\label{tab:KStest_small}
\end{small}
\vskip -0.1in
\end{table}

\subsection{Differences Compared to Autoformer baseline}
Since we use the decomposed encoder-decoder overall architecture as Autoformer, we think it is critical to emphasize the differences. In Autoformer, the authors consider a nice idea to use the top-k sub-sequence correlation (auto-correlation) module instead of point-wise attention, and the Fourier method is applied to improve the efficiency for sub-sequence level similarity computation. In general, Autoformer can be considered as decomposing the sequence into multiple time domain sub-sequences for feature exaction. In contrast, We use frequency transform to decompose the sequence into multiple frequency domain modes to extract the feature. In particular, we do not use a selective approach in sub-sequence selection. Instead, all frequency features are computed from the whole sequence, and this global property makes our model engage better performance for long sequence.


\section{Conclusions}
This paper proposes a frequency enhanced transformer model for long-term series forecasting which achieves state-of-the-art performance and enjoys linear computational complexity and memory cost. We propose an attention mechanism with low-rank approximation in frequency and a mixture of experts decomposition to control the distribution shifting. The proposed frequency enhanced structure decouples the input sequence length and the attention matrix dimension, leading to the linear complexity. Moreover, we theoretically and empirically prove the effectiveness of the adopted random mode selection policy in frequency. Lastly, extensive experiments show that the proposed model achieves the best forecasting performance on six benchmark datasets in comparison with four state-of-the-art algorithms.

\newpage
\bibliography{6_mybib}
\bibliographystyle{icml2021}

\newpage

\appendix

\section{Related Work}
\label{App:related_works}
In this section, an overview of the literature for time series forecasting will be given. The relevant works include traditional times series models (\ref{App:related_work_tradi_time}), deep learning models (\ref{App:related_work_tradi_time}), Transformer-based models (\ref{App:related_work_transformers}), and the Fourier Transform in neural networks (\ref{App:related_work_fft}).

\subsection{Traditional Time Series Models}
\label{App:related_work_tradi_time}
Data-driven time series forecasting helps researchers understand the evolution of the systems without architecting the exact physics law behind them. After decades of renovation, time series models have been well developed and served as the backbone of various projects in numerous application fields. 
The first generation of data-driven methods can date back to 1970. ARIMA~\cite{box_arima2,box_distribution_1970} follows the Markov process and builds an auto-regressive model for recursively sequential forecasting. However, an autoregressive process is not enough to deal with nonlinear and non-stationary sequences. 
With the bloom of deep neural networks in the new century, recurrent neural networks (RNN) was designed especially for tasks involving sequential data. Among the family of RNNs, LSTM~\cite{hochreiter_long_1997_lstm} and GRU~\cite{GRU_cho_et_al_2014} employ gated structure to control the information flow to deal with the gradient vanishing or exploration problem. DeepAR~\cite{DBLP:journals/corr/FlunkertSG17-deepAR} uses a sequential architecture for probabilistic forecasting by incorporating binomial likelihood. Attention based RNN~\cite{dual-state-attention-rnn-qin} uses temporal attention to capture long-range dependencies. However, the recurrent model is not parallelizable and unable to handle long dependencies.
The temporal convolutional network~\cite{Think_globally_act_locally_tcn_time_series_2019} is another family efficient in sequential tasks. However, limited to the reception field of the kernel, the features extracted still stay local and long-term dependencies are hard to grasp.


\subsection{Transformers for Time Series Forecasting}
\label{App:related_work_transformers}
With the innovation of transformers in natural language processing~\cite{attention_is_all_you_need,Bert/NAACL/Jacob} and computer vision tasks~\cite{Transformers-for-image-at-scale/iclr/DosovitskiyB0WZ21,DBLP:Global-filter-FNO-in-cv}, transformer-based models are also discussed, renovated, and applied in time series forecasting~\cite{haoyietal-informer-2021,Autoformer}. In sequence to sequence time series forecasting tasks an encoder-decoder architecture is popularly employed. The self-attention and cross-attention mechanisms are used as the core layers in transformers. 
However, when employing a point-wise connected matrix, the transformers suffer from quadratic computation complexity. 

To get efficient computation without sacrificing too much on performance, the earliest modifications specify the attention matrix with predefined patterns. Examples include: \cite{Block-wise-attention/emnlp/QiuMLYW020} uses block-wise attention which reduces the complexity to the square of block size. Longformer~\cite{Longformer/2020/Iz-Beltagy/corr/abs-2004-05150} employs a stride window with fixed intervals. LogTrans~\cite{Log-transformer-shiyang-2019} uses log-sparse attention and achieves $N\log^2 N$ complexity. H-transformer~\cite{DBLP:conf/acl/H-transformer} uses a hierarchical pattern for sparse approximation of attention matrix with $O(n)$ complexity. Some work uses a combination of patterns (BIGBIRD~\cite{DBLP:conf/nips/Zaheer/BIGBIRD}) mentioned above. Another strategy is to use dynamic patterns: Reformer~\cite{DBLP:conf/iclr/KitaevKL20-reformer} introduces a local-sensitive hashing which reduces the complexity to $N\log N$. 
\cite{DBLP:conf/acl/H-transformer} introduces a hierarchical pattern. Sinkhorn~\cite{sparse-sinkhorn-attention/YiTay/icml/TayBYMJ20} employs a block sorting method to achieve quasi-global attention with only local windows. 

Similarly, some work employs a top-k truncating to accelerate computing: Informer~\cite{haoyietal-informer-2021} uses a KL-divergence based method to select top-k in attention matrix. This sparser matrix costs only $N\log N$ in complexity. Autoformer~\cite{Autoformer} introduces an auto-correlation block in place of canonical attention to get the sub-series level attention, which achieves $N\log N$ complexity with the help of Fast Fourier transform and top-k selection in an auto-correlation matrix. 

Another emerging strategy is to employ a low-rank approximation of the attention matrix. Linformer~\cite{DBLP:journals/corr/abs-2006-04768-linformer} uses trainable linear projection to compress the sequence length and achieves $O(n)$ complexity and theoretically proves the boundary of approximation error based on JL lemma. Luna~\cite{DBLP:journals/corr/LUNA} develops a nested linear structure with $O(n)$ complexity. Nystr\"{o}former~\cite{DBLP:conf/aaai/XiongZCTFLS21/Nystroformer} leverages the idea of Nystr\"{o}m approximation in the attention mechanism and achieves an $O(n)$ complexity. Performer~\cite{DBLP:conf/iclr/Choromanski/Performer} adopts an orthogonal random features approach to efficiently model kernelizable attention mechanisms.

\subsection{Fourier Transform in Transformers}
\label{App:related_work_fft}
Thanks to the algorithm of fast Fourier transform (FFT), the computation complexity of Fourier transform is compressed from $N^2$ to $N\log N$. The Fourier transform has the property that convolution in the time domain is equivalent to multiplication in the frequency domain. Thus the FFT can be used in the acceleration of convolutional networks~\cite{DBLP:MathieuHL13_fft_in_covolution}. FFT can also be used in efficient computing of auto-correlation function, which can be used as a building neural networks block \cite{Autoformer} and also useful in numerous anomaly detection tasks~\cite{IEEE-Homayouni-Autocorrelation-LSTM-anomaly-detection}. 
\cite{Fourier-Neural-Operator,Multiwavelet-based-Operator-Learning} first introduced Fourier Neural Operator in solving partial differential equations (PDEs). FNO is used as an inner block of networks to perform efficient representation learning in the low-frequency domain. FNO is also proved efficient in computer vision tasks~\cite{DBLP:Global-filter-FNO-in-cv}. It also serves as a working horse to build the Wavelet Neural Operator (WNO), which is recently introduced in solving PEDs~\cite{Multiwavelet-based-Operator-Learning}. While FNO keeps the spectrum modes in low frequency, random Fourier method use randomly selected modes. \cite{Random-features-for-large-scale-kernel-machines/NIPS2007_013a006f}  proposes to map the input data to a randomized low-dimensional feature space to accelerate the training of kernel machines. \cite{Sampled-softmax-with-random-Fourier-Features/nips/RawatCYSK19}
proposes the Random Fourier softmax (RF-softmax) method that utilizes the powerful Random Fourier Features to enable more efficient and accurate sampling from an approximate softmax distribution. 

To the best of our knowledge, our proposed method is the first work to achieve fast attention mechanism through low rank approximated transformation in frequency domain for time series forecasting.

\section{Low-rank Approximation of Attention} 
\label{app:Low_rank_approximation}
In this section, we discuss the low-rank approximation of the attention mechanism. First, we present the Restricted Isometry Property (RIP) matrices whose approximate error bound could be theoretically given in \ref{app:low-rank:rip}. Then in \ref{app:low-rank:fourier}, we follow prior work and present how to leverage RIP matrices and attention mechanisms.

If the signal of interest is sparse or compressible on a fixed basis, then it is possible to recover the signal from fewer measurements. \cite{DBLP:journals/corr/abs-2006-04768-linformer,DBLP:conf/aaai/XiongZCTFLS21/Nystroformer} suggest that the attention matrix is low-rank, so the attention matrix can be well approximated if being projected into a subspace where the attention matrix is sparse. For the efficient computation of the attention matrix, how to properly select the basis of the projection yet remains to be an open question. The basis which follows the RIP is a potential candidate.

\subsection{RIP Matrices}
\label{app:low-rank:rip}
The definition of the RIP matrices is:
\begin{definition}
RIP matrices. Let $m < n$ be positive integers, $\Phi$ be a $m \times n$ matrix with real entries, $\delta>0$, and $K<m$ be an integer. We say that $\Phi$ is $(K, \delta)-RIP$, if for every K-sparse vector $x \in \mathbb{R}^n$ we have $(1-\delta)\|x\| \leq \| \Phi x\| \leq (1+\delta)\|x\|$.
\end{definition}
RIP matrices are the matrices that satisfy the restricted isometry property, discovered by D. Donoho, E. Cand\`es and T. Tao in the field of compressed sensing. RIP matrices might be good choices for low-rank approximation because of their good properties. A random matrix has a negligible probability of not satisfying the RIP and many kinds of matrices have proven to be RIP, for example, Gaussian basis, Bernoulli basis, and Fourier basis.

\begin{theorem}
Let $m < n$ be positive integers, $\delta > 0$, and $K=O(\frac{m}{log^4 n})$. Let $\Phi$ be the random matrix defined by one of the following methods:

(Gaussian basis) Let the entries of $\Phi$ be i.i.d. with a normal distribution $N(0, \frac{1}{m})$.

(Bernoulli basis) Let the entries of $\Phi$ be i.i.d. with a Bernoulli distribution taking the values $\pm \frac{1}{\sqrt{m}}$
m, each with 50\% probability.

(Random selected Discrete Fourier basis) Let $A \subset \{0,...,n-1\}$ be a random subset of size m. Let $\Phi$ be the matrix obtained from the Discrete Fourier transform matrix (i.e. the matrix F with entries $F[l,j]=\exp^{-2 \pi ilj/n}/\sqrt{n}$) for $l,j \in \{0,..,n-1\}$ by selecting the rows indexed by A. 

Then $\Phi$ is $(K,\sigma)-RIP$ with probability $p \approx 1-e^{-n}$.
\label{theo:rip_all}
\end{theorem}

Theorem \ref{theo:rip_all} states that Gaussian basis, Bernoulli basis and Fourier basis follow RIP. In the following section, the Fourier basis is used as an example and show how to use RIP basis in low-rank approximation in the attention mechanism.

\subsection{Low-rank Approximation with Fourier Basis/Legendre Polynomials}
\label{app:low-rank:fourier}
Linformer~\cite{DBLP:journals/corr/abs-2006-04768-linformer} demonstrates that the attention mechanism can be approximated by a low-rank matrix. Linformer uses a trainable kernel initialized with Gaussian distribution for the low-rank approximation, While our proposed FEDformer uses Fourier basis/Legendre Polynomials, Gaussian basis, Fourier basis, and Legendre Polynomials all obey RIP, so similar conclusions could be drawn. 

Starting from Johnson-Lindenstrauss lemma~\cite{Johnson1984ExtensionsOL} and using the version from~\cite{robust-concepets-random-projection-arriaga-vempala-2006}, Linformer proves that a low-rank approximation of the attention matrix could be made.

Let $\Phi \in \mathbb{R}^{N \times M}$ be the random selected Fourier basis/Legendre Polynomials. $\Phi$ is RIP matrix. Referring to Theorem \ref{theo:rip_all}, with a probability $p \approx 1-e^{-n}$, for any $x \in \mathbb{R}^N$, we have
\begin{equation}
    (1-\delta)\|x\| \leq \| \Phi x\| \leq (1+\delta)\|x\|.
\end{equation}
Referring to \cite{robust-concepets-random-projection-arriaga-vempala-2006}, with a probability $p \approx 1-4e^{-n}$, for any $x_1,x_2 \in \mathbb{R}^N$, we have
\begin{equation}
   (1-\delta)\|x_1 x_2^\top\| \leq \| x_1\Phi^\top \Phi x_2^\top\| \leq (1+\delta)\|x_1 x_2^\top\|.
\end{equation}
With the above inequation function, we now discuss the case in attention mechanism.
Let the attention matrix $B=softmax(\frac{QK^\top}{\sqrt{d}})=exp(A) \cdot D^{-1}_A$, where $(D_A)_{ii}=\sum^N_{n=1}exp(A_{ni})$. Following Linformer, we can conclude a theorem as (please refer to~\cite{DBLP:journals/corr/abs-2006-04768-linformer} for the detailed proof)
\begin{theorem}
For any row vector $p \in \mathbb{R}^N$ of matrix B and any column vector $v \in \mathbb{R}^N$ of matrix V, with a probability $p = 1 - o(1)$, we have
\begin{equation}
    \| b\Phi^\top \Phi v^\top - bv^\top \| \leq \delta \| bv^\top\|.
\end{equation}
\label{theo:rip-attention-p}
\end{theorem}
Theorem \ref{theo:rip-attention-p} points out the fact that, using Fourier basis/Legendre Polynomials $\Phi$ between the multiplication of attention matrix ($P$) and values ($V$), the computation complexity can be reduced from $O(N^2d)$ to $O(NMd)$, where $d$ is the hidden dimension of the matrix. In the meantime, the error of the low-rank approximation is bounded. However, Theorem \ref{theo:rip-attention-p} only discussed the case which is without the activation function.

Furthermore, with the Cauchy inequality and the fact that the exponential function is Lipchitz continuous in a compact region (please refer to~\cite{DBLP:journals/corr/abs-2006-04768-linformer} for the proof), we can draw the following theorem:
\begin{theorem}
For any row vector $A_i \in \mathbb{R}^N$ in matrix $A$ ($A=\frac{QK^\top}{\sqrt{d}}$), with a probability of $p=1-o(1)$, we have
\begin{equation}
    \| exp(A_i \Phi^\top)  \Phi v^\top - exp(A_i) v^\top \| \leq \delta \| exp(A_i)v^\top\|.
\end{equation}
\label{theo:rip-attention-exp}
\end{theorem}
Theorem \ref{theo:rip-attention-exp} states that with the activation function (softmax), the above discussed bound still holds.

In summary, we can leverage RIP matrices for low-rank approximation of attention. Moreover, there exists theoretical error bound when using a randomly selected Fourier basis for low-rank approximation in the attention mechanism.

\section{Fourier Component Selection}
\label{app:Fourier_component_selection}

Let $X_1(t), \ldots, X_m(t)$ be $m$ time series. By applying Fourier transform to each time series, we turn each $X_i(t)$ into a vector $a_i = (a_{i,1}, \ldots, a_{i,d})^{\top} \in \R^d$. By putting all the Fourier transform vectors into a matrix, we have $A = (a_1, a_2, \ldots, a_m)^{\top} \in \R^{m\times d}$, with each row corresponding to a different time series and each column corresponding to a different Fourier component. 
Here, we propose to select $s$ components from the $d$ Fourier components ($s < d$) uniformly at random. More specifically, we denote by $i_1 < i_2 < \ldots < i_s$ the randomly selected components. We construct matrix $S \in \{0, 1\}^{s\times d}$, with $S_{i, k} = 1$ if $i = i_k$ and $S_{i, k} = 0$ otherwise. Then, our representation of multivariate time series becomes $A' = AS^{\top} \in \R^{m\times s}$. 
The following theorem shows that, although the Fourier basis is randomly selected, under a mild condition, $A'$ can preserve most of the information from $A$. 
\begin{theorem} 
Assume that $\mu(A)$, the coherence measure of matrix $A$, is $\Omega(k/n)$. Then, with a high probability, we have
\[
    |A - P_{A'}(A)| \leq (1 + \epsilon)|A - A_k|
\]
if $s = O(k^2/\epsilon^2)$.
\end{theorem}
\begin{proof}
Following the analysis in Theorem 3 from \cite{Relative-Error-CUR-Matrix-Decompositions}, we have
\begin{equation}
\begin{aligned}
|A - P_{A'}(A)| & \leq  |A - A'(A')^{\dagger}A_k| \\\notag
&= |A - (AS^{\top})(AS^{\top})^{\dagger}A_k| \\\notag
&=  |A - (AS^{\top})(A_kS^{\top})^{\dagger}A_k|.
\end{aligned}
\end{equation}
Using Theorem 5 from \cite{Relative-Error-CUR-Matrix-Decompositions}, we have, with a probability at least $0.7$,
\[
|A - (AS^{\top})(A_kS^{\top})^{\dagger}A_k| \leq (1 + \epsilon)|A - A_k|
\]
if $s = O(k^2/\epsilon^2\times \mu(A)n/k)$. The theorem follows because $\mu(A) = O(k/n)$. 
\end{proof}

\section{Wavelets}
\label{app:wavelets}
In this section, we present some technical background about Wavelet transform which is used in our proposed framework.

\subsection{Continuous Wavelet Transform}
First, let's see how a function $f(t)$ is decomposed into a set of basis functions $\psi_{\mathrm{s}, \tau}(t)$, called the wavelets. It is known as the continuous wavelet transform or $C W T$. More formally it is written as
$$
\gamma(s, \tau)=\int f(t) \Psi_{s, \tau}^{*}(t) d t,
$$
where * denotes complex conjugation. This equation shows the variables $\gamma(s, \tau)$, $s$ and $\tau$ are the new dimensions, scale, and translation after the wavelet transform, respectively.

The wavelets are generated from a single basic wavelet $\Psi(t)$, the so-called mother wavelet, by scaling and translation as
$$\psi_{s, \tau}(t)=\frac{1}{\sqrt{s}} \psi\left(\frac{t-\tau}{s}\right),$$
where $s$ is the scale factor, $\tau$ is the translation factor, and $\sqrt{s}$ is used for energy normalization across the different scales.

\subsection{Discrete Wavelet Transform}
Continues wavelet transform maps a one-dimensional signal to a two-dimensional time-scale joint representation which is highly redundant. To overcome this problem, people introduce discrete wavelet transformation (DWT) with mother wavelet as
$$
\psi_{j, k}(t)=\frac{1}{\sqrt{s_{0}^{j}}} \psi\left(\frac{t-k \tau_{0} s_{0}^{j}}{s_{0}^{j}}\right)
$$
DWT is not continuously scalable and translatable but can be scaled and translated in discrete steps. 
Here $j$ and $k$ are integers and $s_{0}>1$ is a fixed dilation step. The translation factor $\tau_{0}$ depends on the dilation step. The effect of discretizing the wavelet is that the time-scale space is now sampled at discrete intervals. We usually choose $s_{0}=2$ so that the sampling of the frequency axis corresponds to dyadic sampling. For the translation factor, we usually choose $\tau_{0}=1$ so that we also have a dyadic sampling of the time axis.

When discrete wavelets are used to transform a continuous signal, the result will be a series of wavelet coefficients and it is referred to as the wavelet decomposition. 

\subsection{Orthogonal Polynomials}
The next thing we need to focus on is orthogonal polynomials (OPs), which will serve as the mother wavelet function we introduce before. A lot of properties have to be maintained to be a mother wavelet, like admissibility condition, regularity conditions, and vanishing moments. In short, we are interested in the OPs that are non-zero over a finite domain and are zero almost everywhere else. Legendre is a popular set of OPs used it in our work here. Some other popular OPs can also be used here like Chebyshev without much modification. 
\subsection{Legendre Polynomails}
The Legendre polynomials are defined with respect to (w.r.t.) a uniform weight function $w_{L}(x)=1$ for $-1 \leqslant x \leqslant 1$ or $w_{L}(x)=\mathbf{1}_{[-1,1]}(x)$ such that
$$
\int_{-1}^{1} P_{i}(x) P_{j}(x) d x= \begin{cases}\frac{2}{2 i+1} & i=j, \\ 0 & i \neq j .\end{cases}
$$
Here the function is defined over $[-1,1]$, but it can be extended to any interval $[a,b]$ by performing different shift and scale operations.

\subsection{Multiwavelets}
The multiwavelets which we use in this work combine advantages of the wavelet and OPs we introduce before. Other than projecting a given function onto a single wavelet function, multiwavelet projects it onto a subspace of degree-restricted polynomials. In this work, we restricted our exploration to one family of OPs: Legendre Polynomials.

First, the basis is defined as: A set of orthonormal basis w.r.t. measure $\mu$, are $\phi_{0}, \ldots, \phi_{k-1}$ such that $\left\langle\phi_{i}, \phi_{j}\right\rangle_{\mu}=\delta_{i j} .$ With a specific measure (weighting function $w(x)$), the orthonormality condition can be written as $\int \phi_{i}(x) \phi_{j}(x) w(x) d x=\delta_{i j}$.

Follow the derivation in \cite{Multiwavelet-based-Operator-Learning}, through using the tools of Gaussian Quadrature and Gram-Schmidt Orthogonalizaition,
the filter coefficients of multiwavelets using Legendre polynomials can be written as 
$$
\begin{aligned}
H_{i j}^{(0)} &=\sqrt{2} \int_{0}^{1 / 2} \phi_{i}(x) \phi_{j}(2 x) w_{L}(2 x-1) d x \\
&=\frac{1}{\sqrt{2}} \int_{0}^{1} \phi_{i}(x / 2) \phi_{j}(x) d x \\
&=\frac{1}{\sqrt{2}} \sum_{i=1}^{k} \omega_{i} \phi_{i}\left(\frac{x_{i}}{2}\right) \phi_{j}\left(x_{i}\right).
\end{aligned}
$$
For example, if $k=3$, following the formula, the filter coefficients are derived as follows
$$
\resizebox{1.0\hsize}{!}{
\begin{math}
\begin{aligned}
&H^0=[\begin{array}{ccc}
\frac{1}{\sqrt{2}} & 0 & 0 \\
-\frac{\sqrt{3}}{2 \sqrt{2}} & \frac{1}{2 \sqrt{2}} & 0 \\
0 & -\frac{\sqrt{15}}{4 \sqrt{2}} & \frac{1}{4 \sqrt{2}}
\end{array}],H^1=[\begin{array}{ccc}
\frac{1}{\sqrt{2}} & 0 & 0 \\
\frac{\sqrt{3}}{2 \sqrt{2}} & \frac{1}{2 \sqrt{2}} & 0 \\
0 & \frac{\sqrt{15}}{4 \sqrt{2}} & \frac{1}{4 \sqrt{2}}
\end{array}], \\
&G^0=[\begin{array}{ccc}
\frac{1}{2 \sqrt{2}} & \frac{\sqrt{3}}{2 \sqrt{2}} & 0 \\
0 & \frac{1}{4 \sqrt{2}} & \frac{\sqrt{15}}{4 \sqrt{2}} \\
0 & 0 & \frac{1}{\sqrt{2}}
\end{array}],G^1=[\begin{array}{ccc}
-\frac{1}{2 \sqrt{2}} & \frac{\sqrt{3}}{2 \sqrt{2}} & 0 \\
0 & -\frac{1}{4 \sqrt{2}} & \frac{\sqrt{15}}{4 \sqrt{2}} \\
0 & 0 & -\frac{1}{\sqrt{2}}
\end{array}]
\end{aligned}
\end{math}
}
$$

\section{Output Distribution Analysis}
\label{app:output_dis_analy}


\begin{table*}[t!]
\caption{Kolmogrov-Smirnov test P value for long sequence time-series forecasting output on ETT dataset (full experiment)}
\label{tab:sample-table-kstest-large}
\begin{center}
\scalebox{0.9}{
\begin{tabular}{c|c|ccccccc}
\toprule
\multicolumn{2}{c|}{Methods}&Transformer&LogTrans&Informer&Reformer&Autoformer&FEDformer&True\\ 
\midrule

\multirow{4}{*}{\rotatebox{90}{ETTm1}} 
& 96 & 0.0090 & 0.0073 & 0.0055 & 0.0055 &0.020 & \textbf{0.048} & {0.023} \\
& 192 & 0.0052 & 0.0043 & 0.0029 & 0.0013 &0.015 & \textbf{0.028} & {0.013} \\
& 336 & 0.0022 & 0.0026 & 0.0019 & 0.0006 &0.012 & \textbf{0.015} & {0.010} \\
& 720 & 0.0023 & 0.0064 & 0.0016 & 0.0011 &0.008 & \textbf{0.014} & {0.004} \\
\midrule
\multirow{4}{*}{\rotatebox{90}{ETTm2}} 
& 96 & 0.0012 & 0.0025 & 0.0008 & 0.0028& \textbf{0.078} & {0.071} & {0.087} \\
& 192 & 0.0011 & 0.0011 & 0.0006 & 0.0015& \textbf{0.047} & {0.045} & {0.060} \\
& 336 & 0.0005 & 0.0011 & 0.00009 & 0.0007& 0.027 & \textbf{0.028} & {0.042} \\
& 720 & 0.0008 & 0.0005 & 0.0002 & 0.0005& \textbf{0.023} & {0.021} & {0.023} \\

\midrule
\multicolumn{2}{c|}{Count}& 0 & 0 & 0 & 0 & 3 & 5 & NA \\
\bottomrule
\end{tabular}
}
\label{tab:KStest}
\end{center}
\vskip -0.1in
\end{table*}

\subsection{Bad Case Analysis}

Using vanilla Transformer as baseline model, we demonstrate two bad long-term series forecasting cases in ETTm1 dataset as shown in the following Figure \ref{fig_output_dis_analy}.

\begin{figure}[h]
\begin{minipage}{\linewidth/2}
    \centering
    \includegraphics[width=\linewidth]{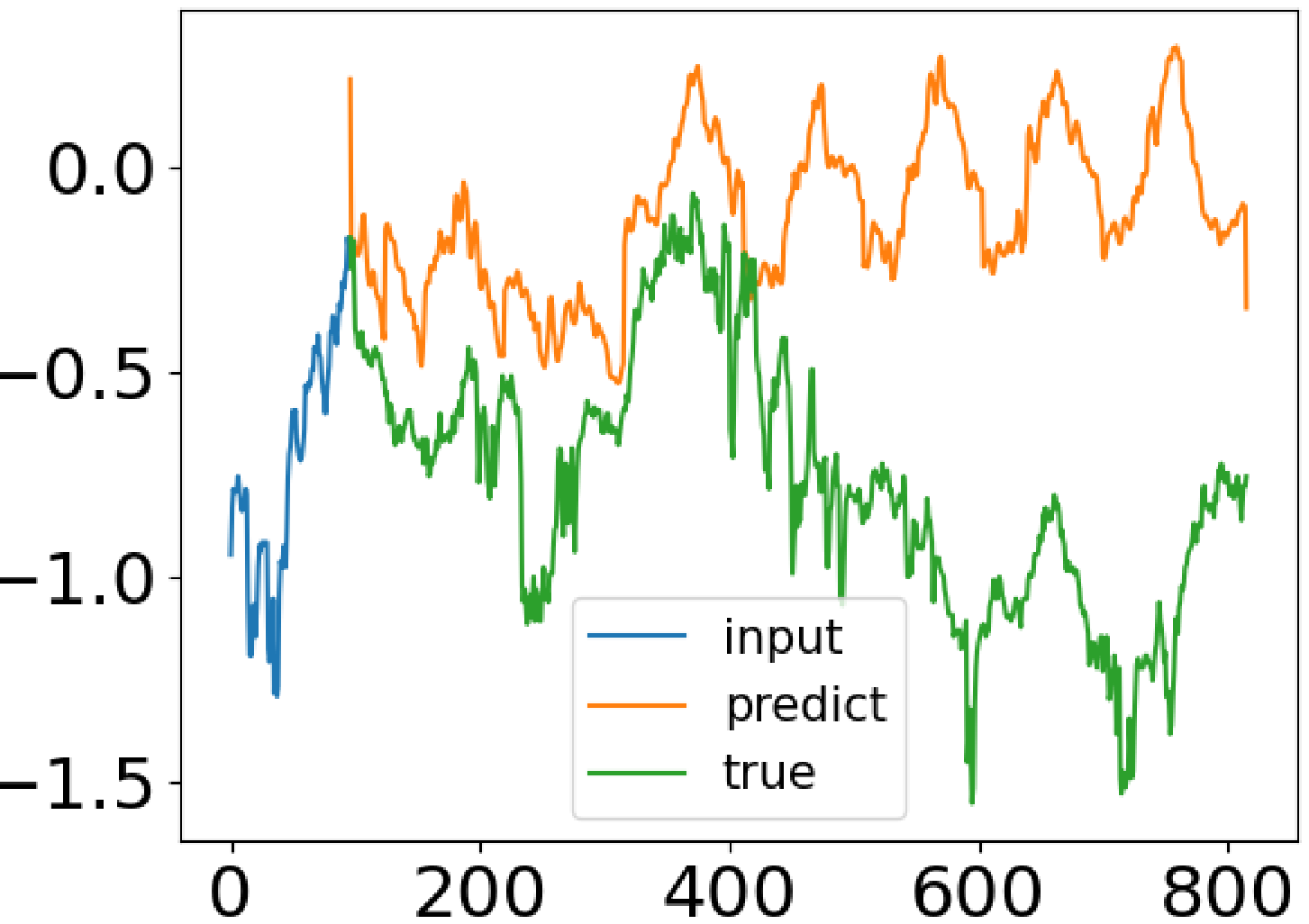}
    \end{minipage}\hfill
    \begin{minipage}{\linewidth/2}
     \centering
    \includegraphics[width=\linewidth]{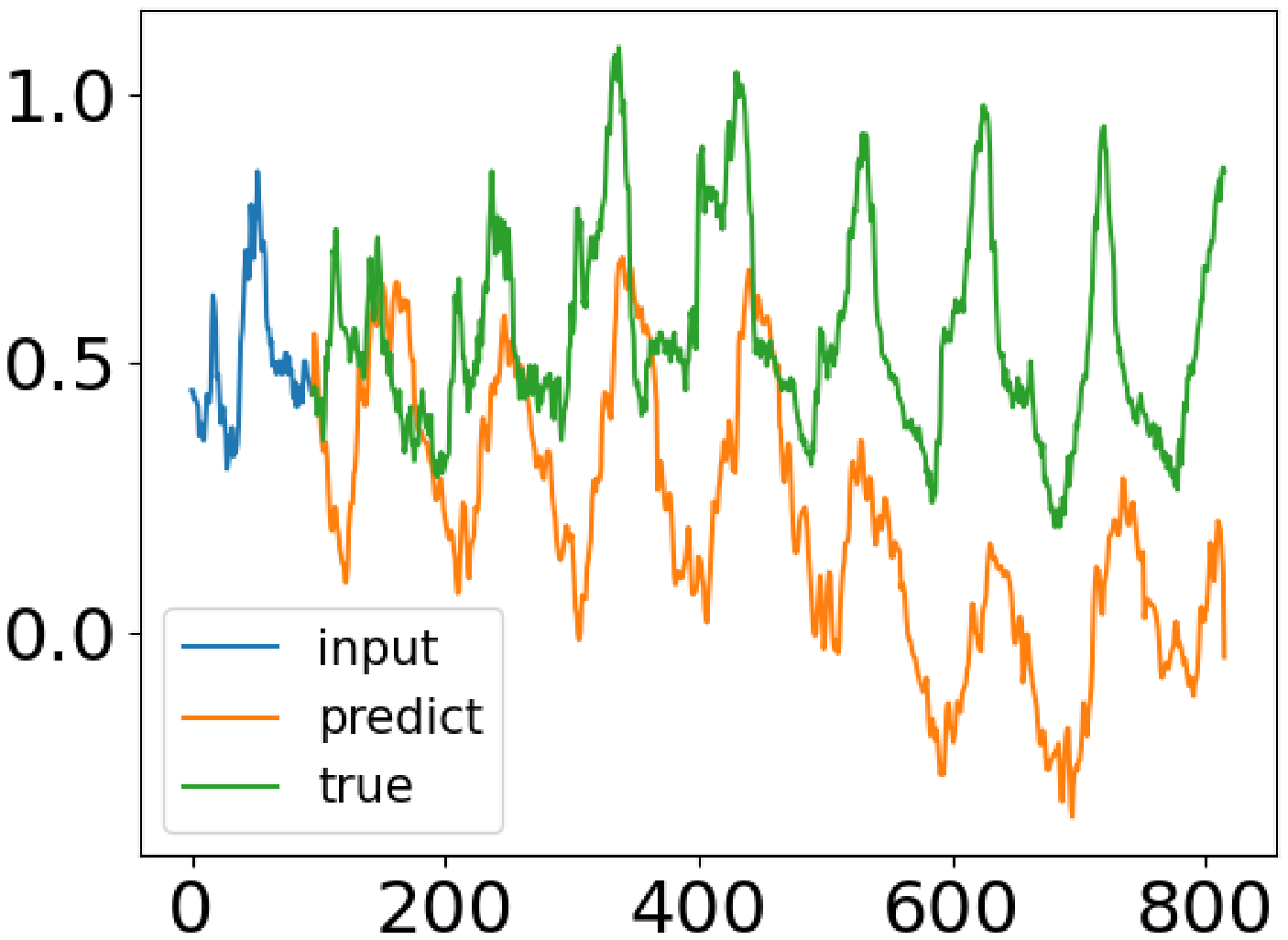}
    \end{minipage}\hfill
    \caption{Different distribution between ground truth and forecasting output from vanilla Transformer in a real-world ETTm1 dataset. Left: frequency mode and trend shift. Right: trend shift.}
    \label{fig_output_dis_analy}
\end{figure}


The forecasting shifts in Figure \ref{fig_output_dis_analy} is particularly related to the point-wise generation mechanism adapted by the vanilla Transformer model. To the contrary of classic models like Autoregressive integrated moving average (ARIMA) which has a predefined data bias structure for output distribution, Transformer-based models forecast each point independently and solely based on the overall MSE loss learning. This would result in different distribution between ground truth and forecasting output in some cases, leading to performance degradation.



\subsection{Kolmogorov-Smirnov Test}
We adopt Kolmogorov-Smirnov (KS) test to check whether the two data samples come from the same distribution. 
KS test is a nonparametric test of the equality of continuous or discontinuous, two-dimensional probability distributions. In essence, the test answers the question ``what is the probability that these two sets of samples were drawn from the same (but unknown) probability distribution". It quantifies a distance between the empirical distribution function of two samples. 
The Kolmogorov-Smirnov statistic is 
$$
D_{n, m}=\sup _{x}\left|F_{1, n}(x)-F_{2, m}(x)\right|
$$
where $F_{1, n}$ and $F_{2, m}$ are the empirical distribution functions of the first and the second sample respectively, and sup is the supremum function.
For large samples, the null hypothesis is rejected at level $\alpha$ if
$$
D_{n, m}>\sqrt{-\frac{1}{2}\ln \left(\frac{\alpha}{2}\right)} \cdot \sqrt{\frac{n+m}{n \cdot m}},
$$
where $n$ and $m$ are the sizes of the first and second samples respectively.

\subsection{Distribution Experiments and Analysis}
Though the KS test omits the temporal information from the input and output sequence, it can be used as a tool to measure the global property of the foretasting output sequence compared to the input sequence. 
The null hypothesis is that the two samples come from the same distribution. 
We can tell that if the P-value of the KS test is large and then the null hypothesis is less likely to be rejected for true output distribution. 

We applied KS test on the output sequence of 96-720 prediction tasks for various models on the ETTm1 and ETTm2 datasets, and the results are summarized in Table \ref{tab:sample-table-kstest-large}. In the test, we compare the fixed 96-time step input sequence distribution with the output sequence distribution of different lengths. 
Using a 0.01 P-value as statistics, various existing Transformer baseline models have much less P-value than 0.01 except Autoformer, which indicates they have a higher probability to be sampled from the different distributions.
Autoformer and FEDformer have much larger P value compared to other models, which mainly contributes to their seasonal trend decomposition mechanism. Though we get close results from ETTm1 by both models, the proposed FEDformer has much larger P-values in ETTm1. And it is the only model whose null hypothesis can not be rejected with P-value larger than 0.01 in all cases of the two datasets, implying that the output sequence generated by FEDformer shares a more similar distribution as the input sequence than others and thus justifies the our design motivation of FEDformer as discussed in Section~\ref{sec_intro}.

Note that in the ETTm1 dataset, the True output sequence has a smaller P-value compared to our FEDformer's predicted output, it shows that the model's close output distribution is achieved through model's control other than merely more accurate prediction. This analysis shed some light on why the seasonal-trend decomposition architecture can give us better performance in long-term forecasting. The design is used to constrain the trend (mean) of the output distribution. Inspired by such observation, we design frequency enhanced block to constrain the seasonality (frequency mode) of the output distribution.


\section{Supplemental Experiments}


\begin{table}[t]
\caption{Summarized feature details of six datasets.}
\label{tab:dataset}
\vskip 0.15in
\begin{center}
\begin{small}
\begin{sc}
\begin{tabular}{l|cccr}
\toprule
Dataset & len & dim & freq \\
\midrule
ETTm2 & 69680 & 8 & 15 min\\
Electricity & 26304 & 322 & 1h & \\
Exchange & 7588 & 9 & 1 day\\
Traffic & 17544 & 863 & 1h & \\
Weather & 52696 & 22 & 10 min & \\
ILI & 966 & 8 & 7 days\\
\bottomrule
\end{tabular}
\end{sc}
\end{small}
\end{center}
\vskip -0.1in
\end{table}

\begin{table*}[t!]
\centering
\caption{Multivariate long sequence time-series forecasting results on ETT full benchmark. The best results are highlighted in bold.}
\vskip 0.05in
\scalebox{0.85}{
\begin{tabular}{c|c|cccccccccccccccc}
\toprule
\multicolumn{2}{c|}{Methods}&\multicolumn{2}{c|}{FEDformer-f}&\multicolumn{2}{c|}{FEDformer-w}&\multicolumn{2}{c|}{Autoformer}&\multicolumn{2}{c|}{Informer}&\multicolumn{2}{c|}{LogTrans}&\multicolumn{2}{c}{Reformer}\\
\midrule
\multicolumn{2}{c|}{Metric} & MSE  & MAE & MSE & MAE& MSE  & MAE& MSE  & MAE& MSE  & MAE& MSE  & MAE\\
\midrule
\multirow{4}{*}{\rotatebox{90}{$ETTh1$}}
& 96  &\textbf{0.376} &\textbf{0.419} &0.395 &0.424 & 0.449& 0.459& 0.865 & 0.713& 0.878& 0.740& 0.837& 0.728\\
& 192 &\textbf{0.420} &\textbf{0.448} &0.469 &0.470 & 0.500& 0.482& 1.008 & 0.792& 1.037& 0.824& 0.923& 0.766\\
& 336 &\textbf{0.459} &\textbf{0.465} &0.530 &0.499 & 0.521& 0.496& 1.107 & 0.809& 1.238& 0.932& 1.097& 0.835\\
& 720 &\textbf{0.506} &\textbf{0.507} &0.598 &0.544 & 0.514& 0.512& 1.181 &0.865&  1.135& 0.852& 1.257& 0.889\\
\midrule
\multirow{4}{*}{\rotatebox{90}{$ETTh2$}}
& 96  &\textbf{0.346} &\textbf{0.388} &0.394 &0.414 & 0.358& 0.397& 3.755& 1.525& 2.116& 1.197 &2.626 &1.317\\
& 192 &\textbf{0.429} &\textbf{0.439} &0.439 &0.445 & 0.456& 0.452& 5.602& 1.931& 4.315& 1.635 &11.12 &2.979\\
& 336 &0.496 &0.487 &\textbf{0.482} &\textbf{0.480} & 0.482& 0.486& 4.721& 1.835& 1.124& 1.604 &9.323 &2.769\\
& 720 &\textbf{0.463} &\textbf{0474} &0.500 &0.509 & 0.515& 0.511& 3.647& 1.625& 3.188& 1.540 &3.874 &1.697\\
\midrule
\multirow{4}{*}{\rotatebox{90}{$ETTm1$}}
& 96  & 0.379 & 0.419 &\textbf{0.378} &\textbf{0.418} & 0.505& 0.475& 0.672& 0.571& 0.600& 0.546 &0.538 &0.528\\
& 192 & \textbf{0.426} & \textbf{0.441} &0.464 &0.463 & 0.553& 0.496& 0.795& 0.669& 0.837& 0.700 &0.658 &0.592\\
& 336 & \textbf{0.445} & \textbf{0.459} &0.508 &0.487 & 0.621& 0.537& 1.212& 0.871& 1.124& 0.832 &0.898 &0.721\\
& 720 & \textbf{0.543} & \textbf{0.490} &0.561 &0.515 & 0.671& 0.561& 1.166& 0.823& 1.153& 0.820 &1.102 &0.841\\
\midrule
\multirow{4}{*}{\rotatebox{90}{$ETTm2$}} &96 & \textbf{0.203} & \textbf{0.287} &0.204 &0.288 &0.255  &0.339  &0.365  &0.453  &0.768  &0.642  &0.658  &0.619 \\
                        & 192 & \textbf{0.269} & \textbf{0.328} &0.316  &0.363  &0.281 &0.340 &0.533  &0.563  &0.989  &0.757  &1.078  &0.827 \\
                        & 336 & \textbf{0.325} & \textbf{0.366} & 0.359 &0.387 &0.339  &0.372  &1.363&0.887  &1.334  &0.872  &1.549  &0.972 \\
                        & 720 & \textbf{0.421} & \textbf{0.415} &0.433 &0.432 &0.422  &0.419  &3.379  &1.338 & 3.048 &1.328  &2.631  &1.242 \\
\bottomrule
\end{tabular}
}
\label{tab:multi-benchmarks-ett}
\vskip -0.1in
\end{table*}

\subsection{Dataset Details}
\label{app:exp:dataset}
In this paragraph, the details of the experiment datasets are summarized as follows: 1) ETT \cite{haoyietal-informer-2021} dataset contains two sub-dataset: ETT1 and ETT2, collected from two electricity transformers at two stations. Each of them has two versions in different resolutions (15min \& 1h). ETT dataset contains multiple series of loads and one series of oil temperatures. 2) Electricity\footnote{https://archive.ics.uci.edu/ml/datasets/ElectricityLoadDiagrams 20112014} dataset contains the electricity consumption of clients with each column corresponding to one client. 3) Exchange \cite{lai2018modeling-exchange-dataset} contains the current exchange of 8 countries. 4) Traffic\footnote{http://pems.dot.ca.gov} dataset contains the occupation rate of freeway system across the State of California. 5) Weather\footnote{https://www.bgc-jena.mpg.de/wetter/} dataset contains 21 meteorological indicators for a range of 1 year in Germany. 6) Illness\footnote{https://gis.cdc.gov/grasp/fluview/fluportaldashboard.html} dataset contains the influenza-like illness patients in the United States. Table \ref{tab:dataset} 
summarizes feature details (Sequence Length: Len, Dimension: Dim, Frequency: Freq) of the six datasets. All datasets are split into the training set, validation set and test set by the ratio of 7:1:2.


\subsection{Implementation Details}

\label{app:exp:implement}
Our model is trained using ADAM \cite{kingma_adam:_2017} optimizer with a learning rate of $1e^{-4}$. The batch size is set to 32. An early stopping counter is employed to stop the training process after three epochs if no loss degradation on the valid set is observed. The mean square error (MSE) and mean absolute error (MAE) are used as metrics. All experiments are repeated 5 times and the mean of the metrics is used in the final results. All the deep learning networks are implemented in PyTorch \cite{NEURIPS2019_9015_pytorch} and trained on NVIDIA V100 32GB GPUs.



\begin{table*}[t]
\centering
\caption{Univariate long sequence time-series forecasting results on ETT full benchmark. The best results are highlighted in bold.}
\vskip 0.05in
\scalebox{0.85}{
\begin{tabular}{c|c|cccccccccccccccc}
\toprule
\multicolumn{2}{c|}{Methods}&\multicolumn{2}{c|}{FEDformer-f}&\multicolumn{2}{c|}{FEDformer-w}&\multicolumn{2}{c|}{Autoformer}&\multicolumn{2}{c|}{Informer}&\multicolumn{2}{c|}{LogTrans}&\multicolumn{2}{c}{Reformer}\\
\midrule
\multicolumn{2}{c|}{Metric} & MSE  & MAE & MSE & MAE& MSE  & MAE& MSE  & MAE& MSE  & MAE& MSE  & MAE\\
\midrule

\multirow{4}{*}{\rotatebox{90}{$ETTh1$}}
&96 &0.079 &0.215 &0.080 &0.214 &\textbf{0.071} &\textbf{0.206}&	0.193&	0.377&	0.283&	0.468&	0.532&	0.569\\
&192 &\textbf{0.104} &\textbf{0.245} &0.105 &0.256 & 0.114&	0.262&	0.217&	0.395&	0.234&	0.409&	0.568&	0.575\\
&336 &0.119 &0.270 &0.120 &0.269 &\textbf{0.107}&\textbf{0.258}&	0.202&	0.381&	0.386&	0.546&	0.635&	0.589\\
&720 &0.142 &0.299 &0.127 &0.280 &\textbf{0.126}&\textbf{0.283}&	0.183&	0.355&	0.475&	0.628&	0.762&	0.666\\
\midrule
\multirow{4}{*}{\rotatebox{90}{$ETTh2$}}
&96 &\textbf{0.128} &\textbf{0.271} &0.156 &0.306 & 0.153&	0.306&	0.213&	0.373&	0.217&	0.379&	1.411&	0.838\\
&192 &\textbf{0.185} &\textbf{0.330} &0.238 &0.380 & 0.204&	0.351&	0.227&	0.387&	0.281&	0.429&	5.658&	1.671\\
&336 &\textbf{0.231} &\textbf{0.378} &0.271 &0.412 & 0.246&	0.389&	0.242&	0.401&	0.293&	0.437&	4.777&	1.582\\
&720 &0.278 &0.420 &0.288 &0.438 & \textbf{0.268}&	\textbf{0.409}&	0.291&	0.439&	0.218&	0.387&	2.042&	1.039\\
\midrule
\multirow{4}{*}{\rotatebox{90}{$ETTm1$}}
&96 &\textbf{0.033} &\textbf{0.140} &0.036 &0.149 & 0.056&	0.183&	0.109&	0.277&	0.049&	0.171&	0.296&	0.355\\
&192 &\textbf{0.058} &\textbf{0.186} &0.069 &0.206 & 0.081&	0.216&	0.151&	0.310&	0.157&	0.317&	0.429&	0.474\\
&336 &0.084 &0.231 &\textbf{0.071} &\textbf{0.209} & 0.076&	0.218&	0.427&	0.591&	0.289&	0.459&	0.585&	0.583\\
&720 &\textbf{0.102} &0.250 &0.105 &\textbf{0.248} & 0.110&	0.267&	0.438&	0.586&	0.430&	0.579&	0.782&	0.730\\
\midrule
\multirow{4}{*}{\rotatebox{90}{$ETTm2$}} 
& 96 & 0.067 & 0.198 &\textbf{0.063} &\textbf{0.189} & 0.065 & 0.189 & 0.088 & 0.225 & 0.075 & 0.208 & 0.076  &0.214   \\
& 192 & \textbf{0.102} & \textbf{0.245} &0.110 &0.252 & 0.118 & 0.256 & 0.132 &0.283 & 0.129 &0.275  & 0.132  & 0.290   \\
& 336 & \textbf{0.130} & \textbf{0.279} &0.147 &0.301 & 0.154 & 0.305  &0.180 &0.336 & 0.154 &0.302  & 0.160  & 0.312    \\
& 720 & \textbf{0.178} & \textbf{0.325} &0.219 &0.368 & 0.182 &0.335  &0.300  &0.435 & 0.160 &0.321  & 0.168  &0.335     \\
\bottomrule
\end{tabular}
}
\label{tab:uni-benchmarks-ett}
\end{table*}

\subsection{ETT Full Benchmark}
\label{app:exp:ett_benchmark}
We present the full-benchmark on the four ETT datasets \cite{haoyietal-informer-2021} in Table \ref{tab:multi-benchmarks-ett} (multivariate forecasting) and Table \ref{tab:uni-benchmarks-ett} (univariate forecasting). The ETTh1 and ETTh2 are recorded hourly while ETTm1 and ETTm2 are recorded every 15 minutes. The time series in ETTh1 and ETTm1 follow the same pattern, and the only difference is the sampling rate, similarly for ETTh2 and ETTm2. On average, our FEDformer yields a \textbf{11.5\%} relative MSE reduction for multivariate forecasting, and a \textbf{9.4\%} reduction for univariate forecasting over the SOTA results from Autoformer.

\begin{figure}[h!]
\centering
\includegraphics[width=1\linewidth]{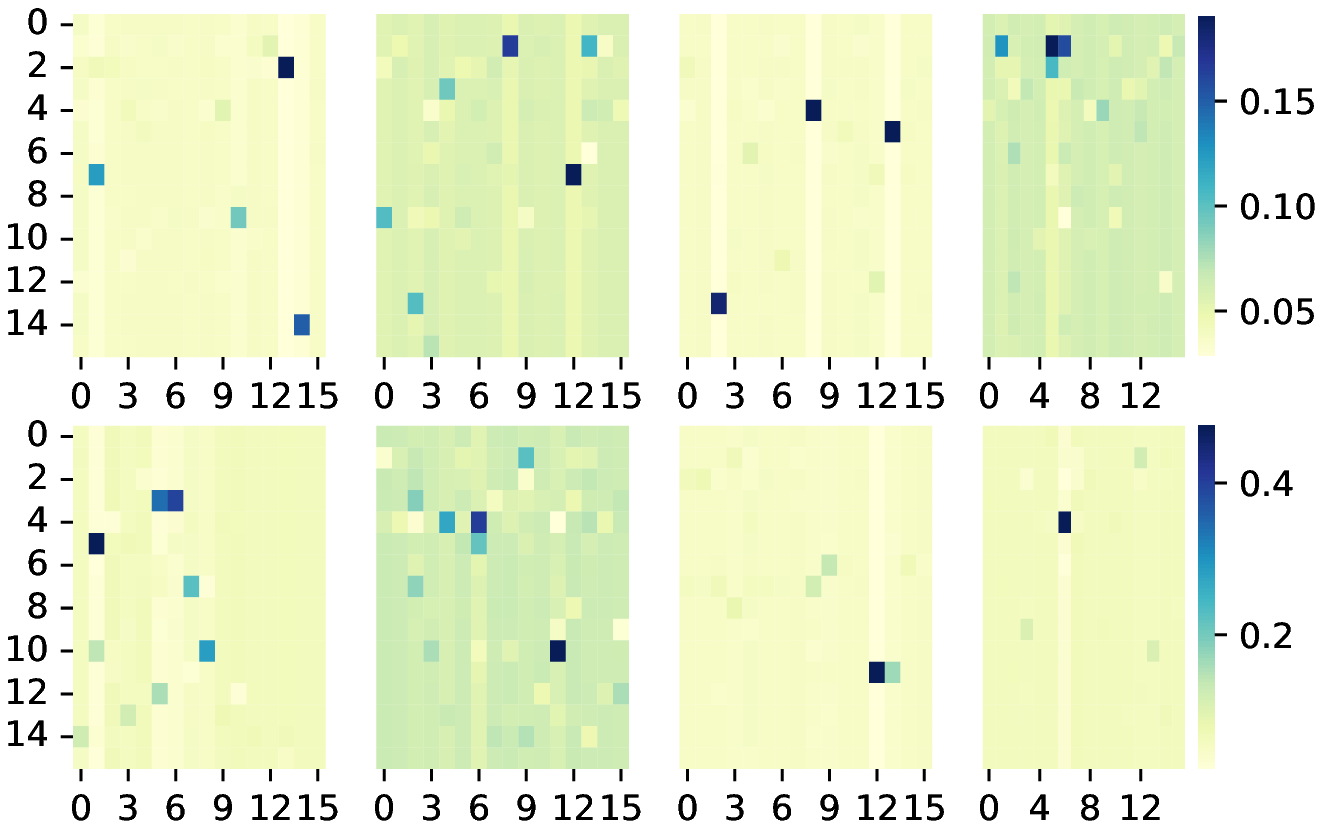}
\includegraphics[width=1\linewidth]{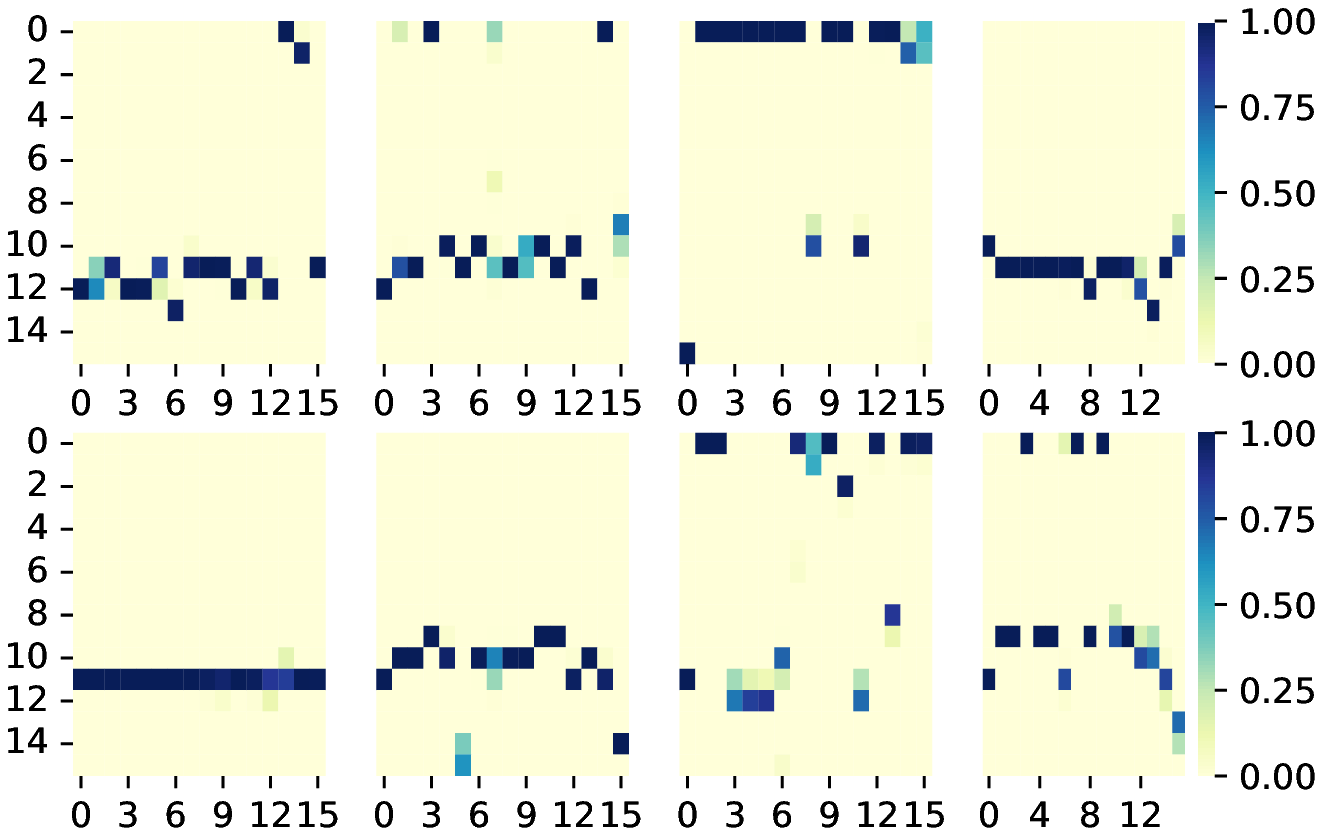}
\caption{Multihead attention map with 8 heads using tanh (top) and softmax (bottom) as activation map for the FEDformer-f training on ETTm2 dataset.}
\label{fig:attention_visulization_0}
\end{figure}

\subsection{Cross Attention Visualization}\label{app:cross_att_visu}

The $\sigma( \tilde{\bm{Q}}\cdot{\tilde{\bm{K}}}^\top )$ can be viewed as the cross attention weight for our proposed frequency enhanced cross attention block. Several different activation functions can be used for attention matrix activation. Tanh and softmax are tested in this work with various performances on different datasets. We use tanh as the default one. 
Different attention patterns are visualized in Figure \ref{fig:attention_visulization_0}. Here two samples of cross attention maps are shown for FEDformer-f training on the ETTm2 dataset using tanh and softmax respectively. It can be seen that attention with Softmax as activation function seems to be more sparse than using tanh. Overall we can see attention in the frequency domain is much sparser compared to the normal attention graph in the time domain, which indicates our proposed attention can represent the signal more compactly. Also this compact representation supports our random mode selection mechanism to achieve linear complexity.


\subsection{Improvements of Mixture of Experts Decomposition}
\label{app:moe_ablation}
We design a mixture of experts decomposition mechanism which adopts a set of average pooling layers to extract the trend and a set of data-dependent weights to combine them. The default average pooling layers contain filters with kernel size 7, 12, 14, 24 and 48 respectively. 
For comparison, we use {single expert} decomposition mechanism which employs a single average pooling layer with a fixed kernel size of 24 as the baseline. 
In Table \ref{tab:moe_ablation}, a comparison study of multivariate forecasting is shown using FEDformer-f model on two typical datasets. It is observed that the designed mixture of experts decomposition brings better performance than the single decomposition scheme.


\begin{table}[t]
\centering
\caption{Performance improvement of the designed mixture of experts decomposition scheme.}
\label{sample-table}
\begin{center}
\begin{small}
\begin{tabular}{c|cc|cc}
\toprule
\multicolumn{1}{c|}{Methods}&\multicolumn{2}{c|}{FEDformer-f}&\multicolumn{2}{c}{FEDformer-f}\\
\midrule
\multicolumn{1}{c|}{Dataset}&\multicolumn{2}{c|}{ETTh1}&\multicolumn{2}{c}{Weather}\\
\midrule
\multicolumn{1}{c|}{Mechanism} & MOE  & Single & MOE &Single\\
\midrule
\multirow{4}{*}{} 
 96 & \textbf{0.217} & 0.238 & 0.376 & \textbf{0.375} \\
 192 & \textbf{0.276} & 0.291 & 0.420 & \textbf{0.412} \\
 336 & \textbf{0.339} & 0.352 &\textbf{ 0.450} & 0.455 \\
 720 & \textbf{0.403} & 0.413 & \textbf{0.496} & 0.502 \\
\midrule
\multicolumn{1}{c|}{Improvement}&\multicolumn{2}{c|}{5.35\%}&\multicolumn{2}{c}{0.57\%} \\
\bottomrule
\end{tabular}
\label{tab:moe_ablation}
\end{small}
\end{center}
\vskip -0.1in
\end{table}

\subsection{Multiple random runs}
Table~\ref{tab:std} lists both mean and standard deviation (STD) for FEDformer-f and Autoformer with 5 runs. We observe a small variance in the performance of FEDformer-f, despite the randomness in frequency selection. 

\vspace{-.2cm}
\begin{table}[h]
\label{sample-table-kstest}
\centering
\vskip -0.2in
\caption{A subset of the benchmark showing both Mean and STD.}
\begin{small}

\scalebox{0.75}{
\begin{tabular}{c|c|ccccc}
\toprule
\multicolumn{2}{c|}{MSE}& ETTm2 & Electricity & Exchange & Traffic\\ 
\midrule
\multirow{4}{*}{\rotatebox{90}{FED-f}} 
& 96 & 0.203 $\pm$ 0.0042 & 0.194 $\pm$ 0.0008 & 0.148 $\pm$ 0.002 & 0.217 $\pm$ 0.008 \\
& 192 & 0.269 $\pm$ 0.0023 & 0.201$\pm$ 0.0015 & 0.270$\pm$ 0.008 & 0.604 $\pm$ 0.004 \\
& 336 & 0.325 $\pm$ 0.0015 & 0.215$\pm$ 0.0018 & 0.460$\pm$ 0.016 & 0.621 $\pm$ 0.006 \\
& 720 & 0.421 $\pm$ 0.0038 & 0.246$\pm$ 0.0020 & 1.195$\pm$ 0.026 & 0.626 $\pm$ 0.003 \\
\midrule
\multirow{4}{*}{\rotatebox{90}{Autoformer}} 
& 96 & 0.255 $\pm$ 0.020 & 0.201$\pm$ 0.003 & 0.197$\pm$ 0.019 & 0.613$\pm$ 0.028 \\
& 192 & 0.281 $\pm$ 0.027 & 0.222$\pm$ 0.003 & 0.300$\pm$ 0.020 & 0.616$\pm$ 0.042 \\
& 336 & 0.339 $\pm$ 0.018 & 0.231$\pm$ 0.006 & 0.509$\pm$ 0.041 & 0.622$\pm$ 0.016 \\
& 720 & 0.422 $\pm$ 0.015 & 0.254$\pm$ 0.007 & 1.447$\pm$ 0.084 & 0.419$\pm$ 0.017 \\
\bottomrule
\end{tabular}
}
\label{tab:std}
\end{small}
\vskip -0.25in
\end{table}

\subsection{Sensitivity to the number of modes: ETTx1 vs ETTx2}
    The choice of modes number depends on data complexity. The time series that exhibits the higher complex patterns requires the larger the number of modes. To verify this claim, we summarize the complexity of ETT datasets, measured by permutation entropy and SVD entropy, in Table~\ref{tab:complexity}. It is observed that ETTx1 has a significantly higher complexity (corresponding to a higher entropy value) than ETTx2, thus requiring a larger number of modes. 
\begin{table}[h]

\vskip -0.27in
\label{sample-table-kstest}
\centering
\begin{small}
\vskip 0.05in
\caption{Complexity experiments for datasets}
\begin{tabular}{c|cccccc}
\toprule
Methods& ETTh1 & ETTh2 & ETTm1 & ETTm2\\ 
\midrule
\multirow{1}{*}{Permutation Entropy} 
&0.954  &0.866  &0.959  &0.788   \\
\midrule
\multirow{1}{*}{SVD Entropy} 
&0.807 &0.495  &0.589  &0.361   \\
\bottomrule
\end{tabular}
\label{tab:complexity}
\end{small}
\vskip -0.28in
\end{table}

\vspace{4mm}

\subsection{When Fourier/Wavelet model performs better} Our high level principle of model deployment is that Fourier-based model is usually better for less complex time series, while wavelet is normally more suitable for complex ones. Specifically, we found that wavelet-based model is more effective on multivariate time series, while Fourier-based one normally achieves better results on univariate time series. As indicated in Table~\ref{tab:uni_multi}, complexity measures on multivariate time series are higher than those on univariate ones. 
\begin{table}[h]

\centering
\caption{Perm Entropy Complexity comparison for multi vs uni}
\begin{small}
\begin{tabular}{c|cccccc}
\toprule
Permutation Entropy& Electricity & Traffic & Exchange & Illness\\ 
\midrule
\multirow{1}{*}{Multivariate} 
&0.910  &0.792  &0.961  &0.960   \\
\midrule
\multirow{1}{*}{Univariate} 
&0.902 &0.790  &0.949  &0.867  \\
\bottomrule
\end{tabular}
\label{tab:uni_multi}
\end{small}
\end{table}







\end{document}